%% file: main.tex
\documentclass[10pt,twocolumn,letterpaper]{article}

\usepackage[pagenumbers]{cvpr} 

\usepackage[accsupp]{axessibility}

\usepackage{titletoc}
\definecolor{cvprblue}{rgb}{0.21,0.49,0.74}
\usepackage[pagebackref,breaklinks,colorlinks,allcolors=cvprblue]{hyperref}

\usepackage{multirow}
\usepackage{float}
\usepackage[most]{tcolorbox}

\usepackage{algorithm}
\usepackage{algpseudocode}
\usepackage{amsmath, mathtools}
\usepackage{makecell}
\usepackage{colortbl}
\usepackage{xcolor}
\usepackage[table]{xcolor} 
\usepackage{wrapfig}

\definecolor{LightOrange}{rgb}{1,0.85,0.8}
\definecolor{blond}{rgb}{0.98, 0.94, 0.75}
\definecolor{blizzardblue}{rgb}{0.67, 0.9, 0.93}
\definecolor{LightGreen}{rgb}{0.93,0.98,0.96}
\definecolor{babypink}{rgb}{0.96, 0.76, 0.76}
\definecolor{classicrose}{rgb}{0.98, 0.8, 0.91}
\definecolor{textboxblue}{RGB}{22, 98, 132}
\definecolor{textboxgrey}{RGB}{252, 252, 252}
\definecolor{commentgreen}{RGB}{25, 107, 36}
\definecolor{lightblue}{rgb}{0.18,0.45,0.71} 

\newcommand{\lgcell}[1]{\cellcolor{green!8}{#1}} 

\def\paperID{****} 
\def\confName{CVPR}
\def\confYear{2026}

\title{MapReduce LoRA: Advancing the Pareto Front in \\Multi-Preference Optimization for Generative Models}

\author{
    Chieh-Yun Chen\textsuperscript{1},
    Zhonghao Wang\textsuperscript{2}$^{\dagger}$,
    Qi Chen\textsuperscript{2},
    Zhifan Ye\textsuperscript{1},
    Min Shi\textsuperscript{1},
    Yue Zhao\textsuperscript{1}, \\
    Yinan Zhao\textsuperscript{2},
    Hui Qu\textsuperscript{2},
    Wei-An Lin\textsuperscript{2},
    Yiru Shen\textsuperscript{2},
    Ajinkya Kale\textsuperscript{2},
    Irfan Essa\textsuperscript{1},
    Humphrey Shi\textsuperscript{1}$^{\dagger}$ \\
{\small \textsuperscript{1}Georgia Tech \textsuperscript{2}Adobe} \\
\centering{{\small \textbf{\texttt{\href{https://github.com/SHI-Labs/MapReduce-LoRA}{\textcolor{magenta}{github.com/SHI-Labs/MapReduce-LoRA}}}}}}
}

\begin{document}

\twocolumn[{%
    \maketitle
        \vspace{-1cm}
        \begin{figure}[H]
        \hsize=\textwidth
        \includegraphics[width=\textwidth]{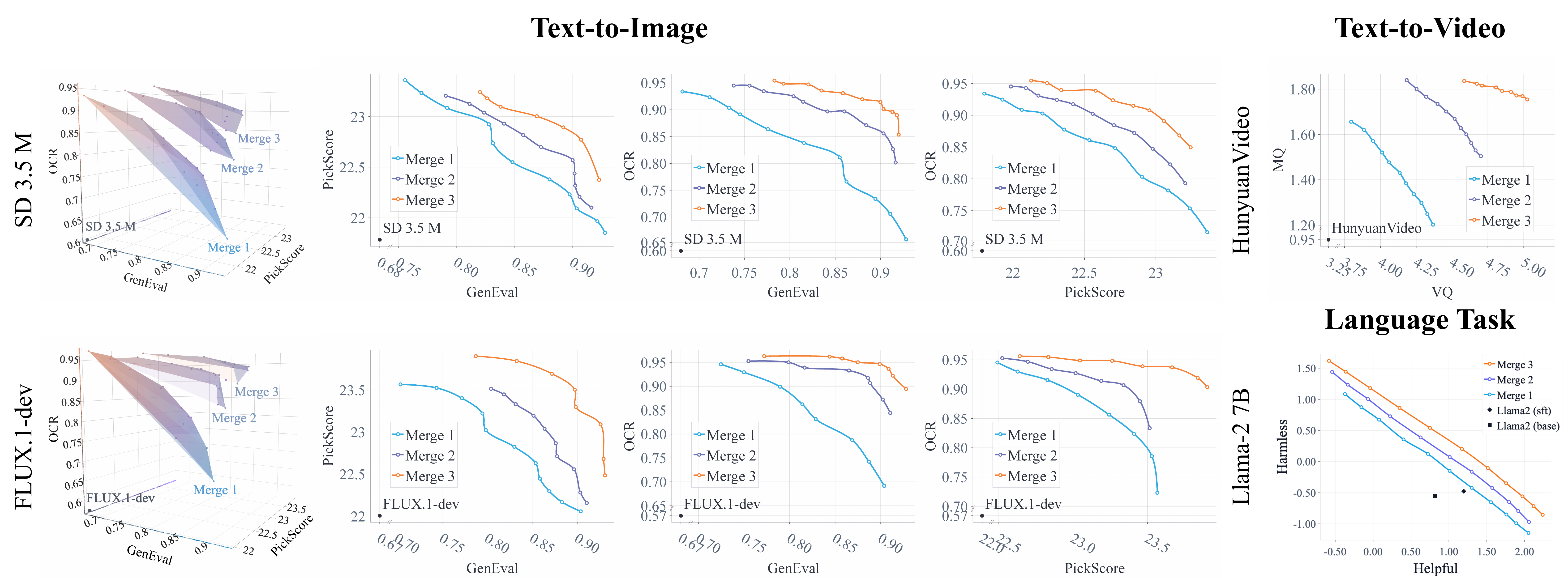}
        \centering
        \vspace{-.3cm}
        \caption{\textbf{MapReduce LoRA advances the Pareto fronts on Text-to-Image, Text-to-Video and language tasks.} Left: On Stable Diffusion 3.5 Medium~\cite{sd35m_hf} (top) and FLUX.1-dev~\cite{flux24} (bottom), we plot the 3D Pareto front over GenEval~\cite{GenEval_NeurIPS'23}, PickScore~\cite{PickScore_NeurIPS'23}, and OCR~\cite{paddleocr_arxiv'25} (Column 1), and the 2D Pareto fronts where one reward weight is set to zero (Columns 2--4). Right (top): On HunyuanVideo~\cite{hunyuanvideo_arxiv'25}, we plot the 2D Pareto front over Visual Quality (VQ) and Motion Quality (MQ). Right (bottom): On Llama-2 7B~\cite{llama2_arxiv23} (following Bone Soup~\cite{bonesoup_acl25}'s setup), we plot the 2D Pareto front on Helpful Assistant task over helpful and harmless rewards. \textit{Please refer to the Supplement~\ref{supp:sec:full_results_fig1} and \ref{supp:sec:language_tasks} for full results.}}
    \label{fig:pareto_front_teaser}
    \end{figure}
}]

\renewcommand{\thefootnote}{\fnsymbol{footnote}}
\footnotetext[2]{Corresponding authors.}
\input{sec/0_abstract}    
\input{sec/1_intro}
\input{sec/2_related_work}
\input{sec/3_preliminary}

\input{sec/4_1_math_proof}

\input{sec/4_2_method_rate}
\input{sec/5_experiment}
\input{sec/6_conclusion}
\input{sec/7_ack}
{
    \small
    \bibliographystyle{ieeenat_fullname}
    \bibliography{main}
}

\clearpage
\onecolumn
\appendix
\input{sec/X_supp}

\end{document}

%% file: sec/0_abstract.tex
\begin{abstract}
    Reinforcement learning from human feedback (RLHF) with reward models has advanced alignment of generative models to human aesthetic and perceptual preferences. However, jointly optimizing multiple rewards often incurs an alignment tax—improving one dimension while degrading others. To address this, we introduce two complementary methods: MapReduce LoRA and Reward-aware Token Embedding (RaTE). MapReduce LoRA trains preference-specific LoRA experts in parallel and iteratively merges them to refine a shared base model; RaTE learns reward-specific token embeddings that compose at inference for flexible preference control. Experiments on Text-to-Image generation (Stable Diffusion 3.5 Medium and FLUX.1-dev) show improvements of 36.1\%, 4.6\%, and 55.7\%, and 32.7\%, 4.3\%, and 67.1\% on GenEval, PickScore, and OCR, respectively. On Text-to-Video generation (HunyuanVideo), visual and motion quality improve by 48.1\% and 90.0\%, respectively. 
    On the language task, Helpful Assistant, with Llama-2 7B, helpful and harmless improve by 43.4\% and 136.7\%, respectively. 
    Our framework sets a new state-of-the-art multi-preference alignment recipe across modalities.
\end{abstract}
    

%% file: sec/1_intro.tex
\section{Introduction}
\label{sec:intro}

\begin{figure*}[t]
    \centering
    \includegraphics[width=1\linewidth]{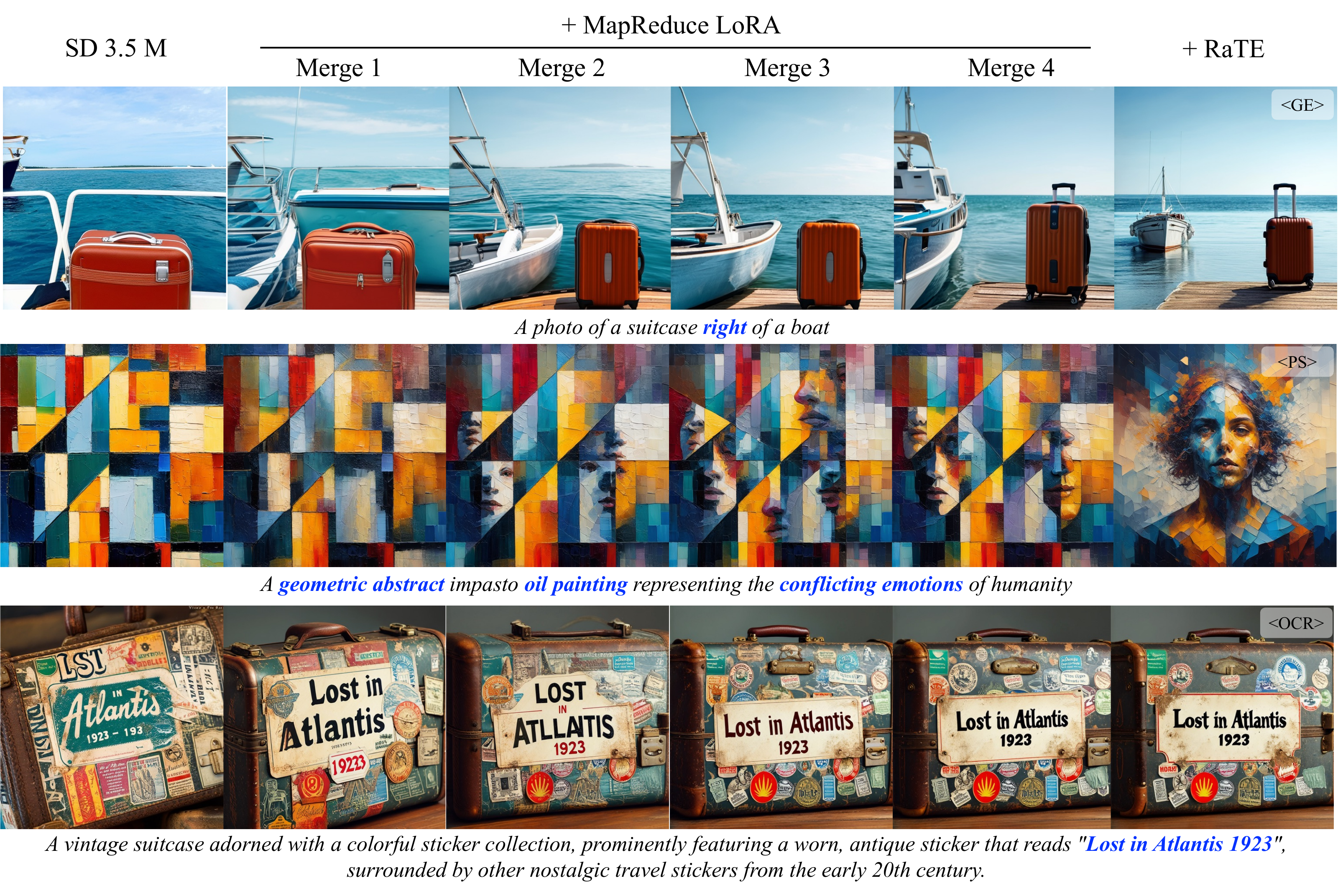}
    \vspace{-.8cm}
     \caption{\textbf{MapReduce LoRA} progressively advances performance across iterations; \textbf{Reward-aware Token Embedding (RaTE)} enables flexible preference control.}
     \label{fig:sd35m_merge_progress}
     \vspace{-.5cm}
  \end{figure*}

Recent progress in Text-to-Image~\cite{sd3_icml24, sd35m_hf, flux24, bagel_arxiv25} and Text-to-Video~\cite{hunyuanvideo_arxiv'25, moviegen_arxiv24, wan_arxiv25} has been driven by flow-based diffusion models~\cite{flow_matching_iclr'23, rf_iclr23}, achieving unprecedented visual fidelity. To better align outputs with human judgment, post-training methods, particularly reinforcement learning from human feedback (RLHF)~\cite{flow_grpo_arxiv'25, dancegrpo_arxiv25, dpo_neurips23, ppo_arxiv17}, have become essential. Reward-based RLHF trains reward models from human annotations to capture specific evaluation criteria, and optimizes the generator by sampling generations, scoring them with the reward models, and updating parameters with reward-weighted objectives. However, human perception of quality is inherently multi-dimensional. In practice, multiple reward models are used to reflect criteria such as text–image alignment~\cite{GenEval_NeurIPS'23, VQAScore_ECCV24}, aesthetic quality~\cite{VILA_CVPR'23}, text rendering~\cite{paddleocr_arxiv'25}, and overall preference~\cite{PickScore_NeurIPS'23, HPSv3_arxiv'25}. This raises a central challenge: how to optimize generative models that can improve across multiple preferences simultaneously without sacrificing any single dimension.

While post-training methods~\cite{flow_grpo_arxiv'25, dancegrpo_arxiv25} perform well under a single reward, they either degrade on unoptimized metrics at evaluation time or suffer competing gradients when jointly optimizing multiple rewards—two manifestations of the alignment tax.
%
Multi-objective reinforcement learning (MORL)~\cite{CaPO_CVPR'25, MOPO_arxiv25, reward_soup_nips'23, selma_neurips24} attempts to optimize multiple rewards simultaneously by combining rewards through weighted mixtures or approximating a Pareto set with test-time weight control. However, weighted mixtures are dominated by easily optimized objectives, leaving harder preferences under-trained or even regressed. Calibrated Preference Optimization (CaPO)~\cite{CaPO_CVPR'25} partially addresses this imbalance but remains limited by fixed weighting and modest performance gains. 
Rewarded Soup~\cite{reward_soup_nips'23} offers a posteriori weight selection but still lags behind models fine-tuned individually per reward.
Consequently, existing approaches struggle to deliver models that generalize well across diverse reward signals. Achieving a unified framework that improves multiple human-aligned preferences efficiently and robustly remains a fundamental challenge in multi-objective post-training.

To address these limitations, we propose two complementary approaches—MapReduce LoRA and Reward-aware Token Embedding (RaTE)—that jointly advance multi-preference alignment. MapReduce LoRA decomposes multi-objective optimization into i) a Map phase that trains reward-specific LoRA experts in parallel, and ii) a Reduce phase that merges experts with user-controlled interpolation, folds the merged adapter into the base, and iterates to advance the Pareto front. Complementarily, RaTE introduces a lightweight inference-time control by assigning each reward a learned token embedding, enabling composable conditioning by appending multiple tokens to the input prompt. Together, MapReduce LoRA and RaTE enable a posteriori customization without retraining, yielding unified generative models that excel across multiple reward dimensions.

Extensive experiments demonstrate that our proposed methods substantially advance the Pareto front of multi-preference alignment across both Text-to-Image, Text-to-Video generation, and Language Tasks (Fig.~\ref{fig:pareto_front_teaser}). On Stable Diffusion 3.5 Medium~\cite{sd35m_hf} and FLUX.1-dev~\cite{flux24}, our method improves GenEval~\cite{GenEval_NeurIPS'23}, PickScore~\cite{PickScore_NeurIPS'23}, and OCR~\cite{paddleocr_arxiv'25} by 36.1\%, 4.6\%, 55.7\% and 32.7\%, 4.3\%, 67.1\%, respectively. Beyond in-domain metrics, untargeted rewards—\ie, VQAScore~\cite{VQAScore_ECCV24}, MPS~\cite{MPS_CVPR'24}, and VILA~\cite{VILA_CVPR'23}—also improve by 1.85\%, 6.49\%, and 19.96\%, validating the robustness and scalability of our method. On HunyuanVideo~\cite{hunyuanvideo_arxiv'25}, visual and motion quality improve by 48.1\% and 90.0\%, achieving state-of-the-art results among post-trained diffusion systems. 
On Llama-2 7B~\cite{llama2_arxiv23}, helpful and harmless improve by 43.4\% and 136.7\%, respectively. 
Collectively, these results highlight the effectiveness of combining MapReduce LoRA and RaTE for efficient, controllable, and human-aligned generation. Our contributions are fourfold:

\begin{itemize}
    \item We introduce MapReduce LoRA, a scalable multi-reward training framework that iteratively advances the Pareto front across preferences.
    \item We propose Reward-aware Token Embedding for flexible, composable inference-time control of reward trade-offs.
    \item MapReduce LoRA achieves state-of-the-art performance on Text-to-Image, Text-to-Video and language tasks, with substantial multi-rewards gains.
    \item MapReduce LoRA exhibits strong generalization to untargeted rewards, demonstrating robust cross-preference alignment.
\end{itemize}

%% file: sec/2_related_work.tex
\section{Related Work}
\label{sec:related}

\subsection{Flow-based Generative Models} Flow Matching (FM)~\cite{flow_matching_iclr'23} trains continuous normalizing flows by regressing the conditional velocity field along a path between the data distribution and a standard normal, yielding a score-free, simulation-free objective with more stable and efficient optimization than diffusion losses. 
Rectified Flow (RF)~\cite{rf_iclr23} specializes FM to a straight-line, approximately constant-speed path, rectifying trajectories and simplifying supervision, which improves stability and sampling efficiency. 
Recent Text-to-Image~\cite{sd3_icml24, sd35m_hf, flux24, bagel_arxiv25, Chen_2025_ICCV} and Text-to-Video~\cite{hunyuanvideo_arxiv'25,moviegen_arxiv24,wan_arxiv25} methods adopt FM/RF to learn the velocity field directly, improving efficiency over diffusion-style denoising and often yielding stronger gradients for conditional generation.

\subsection{Reinforcement Learning from Human Feedback (RLHF)}
RLHF~\cite{rlhf_neurips17,atari_rlhf_neurIPS18} was first used to train agents in simulators and Atari, and later to fine-tune language models for summarization~\cite{shf_arxiv20}. InstructGPT~\cite{instructgpt_neurips22} scaled RLHF to align broad language tasks with a three-stage pipeline: supervised fine-tuning (SFT), reward-model training, and policy optimization with Proximal Policy Optimization (PPO)~\cite{ppo_arxiv17}. Direct Preference Optimization (DPO)~\cite{dpo_neurips23} removes the reward model and online RL, directly optimizing the policy from human preference pairs. Group Relative Policy Optimization (GRPO)~\cite{grpo_arxiv'24} is a PPO-style preference method that drops the critique and uses relative scores across multiple samples per prompt, improving stability and sample efficiency. To apply RL to diffusion models, Denoising Diffusion Policy Optimization (DDPO)~\cite{ddpo_iclr24} formulates the denoising process as a Markov Decision Process (MDP). Building on this MDP formulation, Flow-GRPO~\cite{flow_grpo_arxiv'25} and DanceGRPO~\cite{dancegrpo_arxiv25} extend GRPO~\cite{grpo_arxiv'24} to flow-based generative models by using an ODE-to-SDE strategy to inject stochasticity and overcome the determinism of standard flow models. 

However, single-reward RLHF optimizes one preference dimension in isolation, while generative quality spans multiple and often conflicting objectives. This creates a natural bridge to \textbf{Multi-Objective RL (MORL)}, where the objective is to balance conflicting rewards and approximate a Pareto-optimal solution—precisely the challenge in post-training generative models. 
MORL methods can be categorized into two classes: \textit{a priori}~\cite{CaPO_CVPR'25, MOPO_arxiv25}, which scalarizes objectives into a single signal (\eg, weighted sums), and \textit{a posteriori}~\cite{reward_soup_nips'23, selma_neurips24}, which approximate a Pareto set and enable test-time control. 
CaPO~\cite{CaPO_CVPR'25} calibrates and balances multiple T2I preferences, and MOPO~\cite{MOPO_arxiv25} frames alignment as constrained, KL-regularized optimization; both are \textit{a priori} and lack test-time control. Rewarded Soups~\cite{reward_soup_nips'23} extends Linear Mode Connectivity~\cite{lmc_icml'20, transfer_neurips'20} to RL, showing linear weight soups can approximate Pareto fronts and reduce reward misspecification, yet multi-reward methods often underperform specialized single-reward experts. We therefore propose two \textit{a posteriori} solutions, RaTE and MapReduce LoRA, to address reward conflict. 
Specifically, the iterative merging in MapReduce LoRA shares a technical spirit with periodic averaging in FedAvg~\cite{fedavg_pami23} and DiLoCo~\cite{diloco_arxiv23}; however, while they focus on communication efficiency across GPU clusters, our work leverages this mechanism to navigate the Pareto front for multi-preference alignment.

%% file: sec/3_preliminary.tex
\section{Method}
\label{sec:method}

\subsection{Preliminaries}
\label{sec:prelim}

\begin{figure*}[t]
    \centering
    \includegraphics[width=1.0\linewidth]{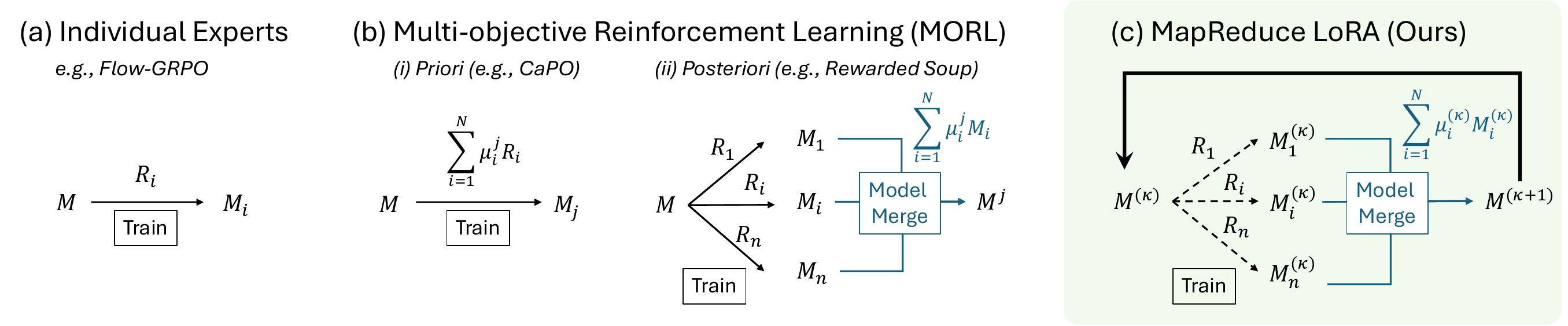}
    \vspace{-.5cm}
     \caption{\textbf{Overview of MapReduce LoRA and comparison with (a) individual experts, \eg, Flow-GRPO~\cite{flow_grpo_arxiv'25}, and (b) Multi-Objective Reinforcement Learning, \eg, CaPO~\cite{CaPO_CVPR'25} and Rewarded soup~\cite{reward_soup_nips'23}.} All methods begin from a base model $M$ and optimize with respect to reward $R$. Our proposed MapReduce LoRA iteratively trains per-reward LoRA experts and initializes iteration $k+1$ using the merged model from iteration $k$. Notably, the black dashed curve for MapReduce LoRA indicates fewer training steps compared with the black solid curves representing other methods.}  
     
     \label{fig:arc}
  \end{figure*}

\noindent\textbf{Low-Rank Adaptation (LoRA)}~\cite{lora_iclr'22} is a leading parameter-efficient fine-tuning method that adapts large models by learning a low-rank update to frozen pre-trained weights. For a linear layer with weight \(W \in \mathbb{R}^{d \times k}\) and input \(x\), the adapted layer is
\[
\big(W + \Delta W\big) x \,=\, W x \, + \, B A x, \quad \text{with} \; \Delta W = B A,
\]
where \(A \in \mathbb{R}^{r \times k}\), \(B \in \mathbb{R}^{d \times r}\), and \(r \ll \min(d,k)\). During fine-tuning, \(W\) remains fixed while only \(A\) and \(B\) are trained, optionally scaled by a factor \(\alpha / r\); at inference, \(\Delta W\) can be merged into \(W\). 
Recent work~\cite{schulman2025lora} finds that, in common post-training settings (small/medium-scale SFT, reasoning, or RL), LoRA matches full fine-tuning performance when rank and layer coverage suffice.
LoRA has been used for style~\cite{multilora_tmlr24} and skill control~\cite{selma_neurips24}, but multi-preference control remains underexplored; we therefore study LoRA merging for multi-preference optimization.
%

\vspace{.2cm}

\noindent\textbf{Textual Inversion}~\cite{ti_iclr'23} learns a new embedding vector for a pseudo-token that represents a specific concept. Given only three to five reference images, it optimizes a single token embedding to encode both the high-level semantics and the fine-grained visual details of the concept, enabling personalized and reference-guided generation without modifying the base model.

\vspace{.2cm}

\noindent\textbf{Pareto Fronts (PF)}~\cite{paretofront_1896} represents the collection of non-dominated solutions, model parameters for which no other candidate can improve one objective without degrading another. Formally,
$\mathrm{PF} = \left\{ \theta \mid \nexists\, \theta' \in \Theta,\; \{ R_i(\theta') \}_i \succ_{\mathbb{R}^N} \{ R_i(\theta) \}_i \right\},$
where $\succ_{\mathbb{R}^N}$ denotes the dominance relation in the N-dimensional reward space indexed by $i$, and $N$ denotes the total number of reward objectives.
\vspace{.2cm}

\noindent\textbf{Group Relative Policy Optimization (GRPO)}~\cite{grpo_arxiv'24, flow_grpo_arxiv'25,dancegrpo_arxiv25} 
is a PPO-style preference optimization algorithm but based on group-normalized rewards. For each prompt $p$, GRPO samples \(G\) results $\{y^{g}\}_{g=1}^{G}$ and computes group-normalized advantages $\hat{A}^{g}$ by z-scoring rewards within that group as shown in Eqn.~(\ref{eqn:advatage}). The policy is then updated with a clipped likelihood-ratio objective and regularized by a KL penalty w.r.t a frozen reference policy as shown in Eqn.~(\ref{eqn:grpo_objective}). 

\footnotesize
\begin{equation}
    \label{eqn:grpo_objective}
  \begin{aligned}
    \mathcal{J}_{\text{GRPO}}
    &= \mathbb{E}_{p}\!\left[
    \frac{1}{G}\sum_{g=1}^{G}\frac{1}{T}\sum_{t=1}^{T}
    \min\big(r_{t}^{g}\,\hat{A}^{g},\,\operatorname{clip}(r_{t}^{g},1-\epsilon,1+\epsilon)\,\hat{A}^{g}\big)
    \right] \\
    &\quad - \beta\,D_{KL}\!\big(\pi_{\theta}(\cdot\mid p)\,\big\|\,\pi_{\text{ref}}(\cdot\mid p)\big),
  \end{aligned}
\end{equation}
\begin{equation}
  \label{eqn:advatage}
  \hat{A}^{g} = \frac{R(y^{g}, p) - \operatorname{mean}\!\big[R(y^{g}, p)\big]_{g=1}^{G}}{\operatorname{std}\!\big[R(y^{g}, p)\big]_{g=1}^{G}}.
\end{equation}

\normalsize

%% file: sec/4_1_math_proof.tex
\subsection{MapReduce LoRA}
\label{sec:mapreduce_lora}
In this section, we first demonstrate that MapReduce is a progressive souping process via averaged proximal consensus optimization. Then, we prove that MapReduce progressively converges toward the joint optimum of the conditions set by multiple reward models. Finally, we explain why MapReduce outperforms the one-shot soup method~\cite{reward_soup_nips'23}. Fig.~\ref{fig:arc} shows the overview of MapReduce LoRA and a comparison with individual experts and multi-objective reinforcement learning~\cite{CaPO_CVPR'25, reward_soup_nips'23}. 

Let $\{f_i(\theta)\}_{i=1}^n$ denote differentiable reward objectives, each derived from a distinct reward model used in GRPO~\cite{grpo_arxiv'24} fine-tuning. The joint optimization goal is to maximize their average:
\begin{equation}
\label{eqn:optimization_fn}
    F(\theta) = \frac{1}{n} \sum_{i=1}^{n} f_i(\theta),
\end{equation}
where $\theta \in \mathbb{R}^d$ are model parameters of the diffusion model. 

MapReduce is implemented by iteratively optimizing multiple reward functions for a few steps and averaging the resulting weights, a process we formalize as progressive souping. At iteration $k$:
\begin{equation}
\begin{aligned}
\label{eqn:progressive_souping}
\theta_i^k = \mathrm{prox}_{\eta f_i} (\theta^k) &= \arg \max_{\theta} (f_i (\theta) - \frac{1}{2 \eta} \| \theta - \theta^k \|^2), \\
  \theta^{k+1} &= \frac{1}{n} \sum_{i=1}^{n} \theta_i^k,
\end{aligned}
\end{equation}
where the proximal term $\frac{1}{2 \eta} \| \theta - \theta^k \|^2$ corresponds to a trust region in GRPO. The overall operator can be written as:
\begin{equation}
\label{eqn:overall_operator}
    T(\theta) = \frac{1}{n} \sum_{i=1}^{n} \mathrm{prox}_{\eta f_i} (\theta).
\end{equation}
Thus, progressive souping performs repeated applications of the averaged proximal map $\theta^{k+1}=T(\theta^k)$, while a one-shot “final soup” corresponds to a single application $T(\theta^0)$.

Now we prove that MapReduce progressively reaches the optimum of Eqn.~(\ref{eqn:optimization_fn}). Because the sampling distribution $p_\theta$ of a pretrained diffusion model changes smoothly with $\theta$ (the score function $\nabla_\theta \log (p_\theta)$ is continuous), we can safely assume that each reward objective $f_i(\theta)$ is locally $\mathcal{L}$-smooth and the aggregated objective Eqn.~(\ref{eqn:optimization_fn}) satisfies the Polyak–Łojasiewicz (PL) condition in a neighborhood containing the optimization trajectory:
\begin{equation}
\label{eqn:pl_condition}
    \frac{1}{2}\|\nabla F(\theta) \|^2 \geq \mu \|F(\theta) - F^*\|,
\end{equation}
for some constant $\mu > 0$. Then, for step size $0 < \eta \leq 1 / \mathcal{L}$, each proximal operator $\mathrm{prox}_{\eta f_i} (\theta)$ is nonexpansive in a local neighborhood, and the overall operator $T(\theta)$ is thus an $\alpha$-averaged operator. The fixed points of $T(\theta)$ coincide with the stationary points of the aggregated objective $\nabla F(\theta^*) = 0$. The progressive-soup iteration in Eqn.~(\ref{eqn:progressive_souping}) converges to a stationary point of Eqn.~(\ref{eqn:optimization_fn}) and satisfies the geometric contraction bound:
\begin{equation}
\label{eqn:geometric_bound}
    \|F(\theta^{k+1}) - F^* \| \leq (1 - c \eta \mu) \| F(\theta^k) - F^* \|,
\end{equation}
for some constant $c \in (0, 1)$.

A one-shot final soup performs a single application of $T$; progressive souping applies $T$ repeatedly, resulting in a smaller sub-optimality gap:
\begin{equation}
\label{eqn:suboptimality}
    \|F(\theta^{m}) - F^* \| \leq (1 - c \eta \mu)^{m} \| F(\theta^0) - F^* \|,
\end{equation}
where $m$ denotes the iteration. Hence, each progressive-soup iteration further contracts toward the joint optimum, while the one-shot soup remains suboptimal unless initialized near a stationary point. The comparison between progressive souping and one-shot soup is shown in Figs.~\ref{fig:pareto_front_teaser}, \ref{fig:sd35m_merge_progress} and \ref{fig:sd35_flux_qual}. \textit{Please refer to the Supplement~\ref{supp:sec:pseudocode} for the pseudocode.}

%% file: sec/4_2_method_rate.tex
\subsection{Reward-aware Token Embedding (RaTE)}
\label{sec:rate}
We propose RaTE to make the token embedding sensitive to user preferences and dynamically control inference-time reward influences. Inspired by Textual Inversion~\cite{ti_iclr'23}, we distill each preference into a single, trainable special token embedding via supervised fine-tuning. Both teacher and student share the same frozen Transformer backbone $M$. We detail the setup below.

\begin{itemize}
    \item\textbf{Teacher:} For each preference $i$, we attach its expert LoRA adapter $\theta_i$ to $M$. This teacher produces the target latent $z_{0,i}^{\text{teacher}}$ for the $i$-th preference.

    \item\textbf{Student:} The student is the frozen $M$ without adapters. Its prompt includes the $i$-th special token; the token embedding $\theta_{\text{token}_i}$ is the only trainable parameter.
\end{itemize}
We distill $\theta_i$ into $\theta_{\text{token}_i}$ using a Flow Matching objective~\cite{flow_matching_iclr'23}. Sample $t$ via a flow-matching timestep schedule (e.g., logit-normal or uniform (depending on the scheduler/model)), sample $\epsilon \sim \mathcal{N}(0,I)$, and set $z_t = (1 - \sigma_t)\, z_{0,i}^{\text{teacher}} + \sigma_t\, \epsilon$, where $\sigma_t$ is given by the scheduler. The student $M$ predicts the velocity from $z_t$ and the conditioned prompt $c(p, \theta_{\text{token}_i})$. The target velocity is $v_{\text{target}} = \epsilon - z_{0,i}^{\text{teacher}}$.
The loss minimizes the Mean Squared Error (MSE) between prediction and target, updating only $\theta_{\text{token}_i}$:

\small
\begin{equation}
\mathcal{L}(\theta_{\text{token}_i}) 
= \mathbb{E}_{p,\, z_{0,i}^{\text{teacher}},\, \epsilon,\, t}
\Big[\big\| M(z_t, t, c(p, \theta_{\text{token}_i})) - v_{\text{target}} \big\|_2^2 \Big].
\end{equation}

\normalsize
After training, we obtain the expert-infused token embedding $\theta_{\text{token}_i}$. At inference, appending its special token to a prompt injects the desired preference. RaTE is lightweight and modular, enabling composable control with other customizations such as LoRA adapters.

%% file: sec/5_experiment.tex
\begin{figure*}[t]
    \centering
    \includegraphics[width=1.0\linewidth]{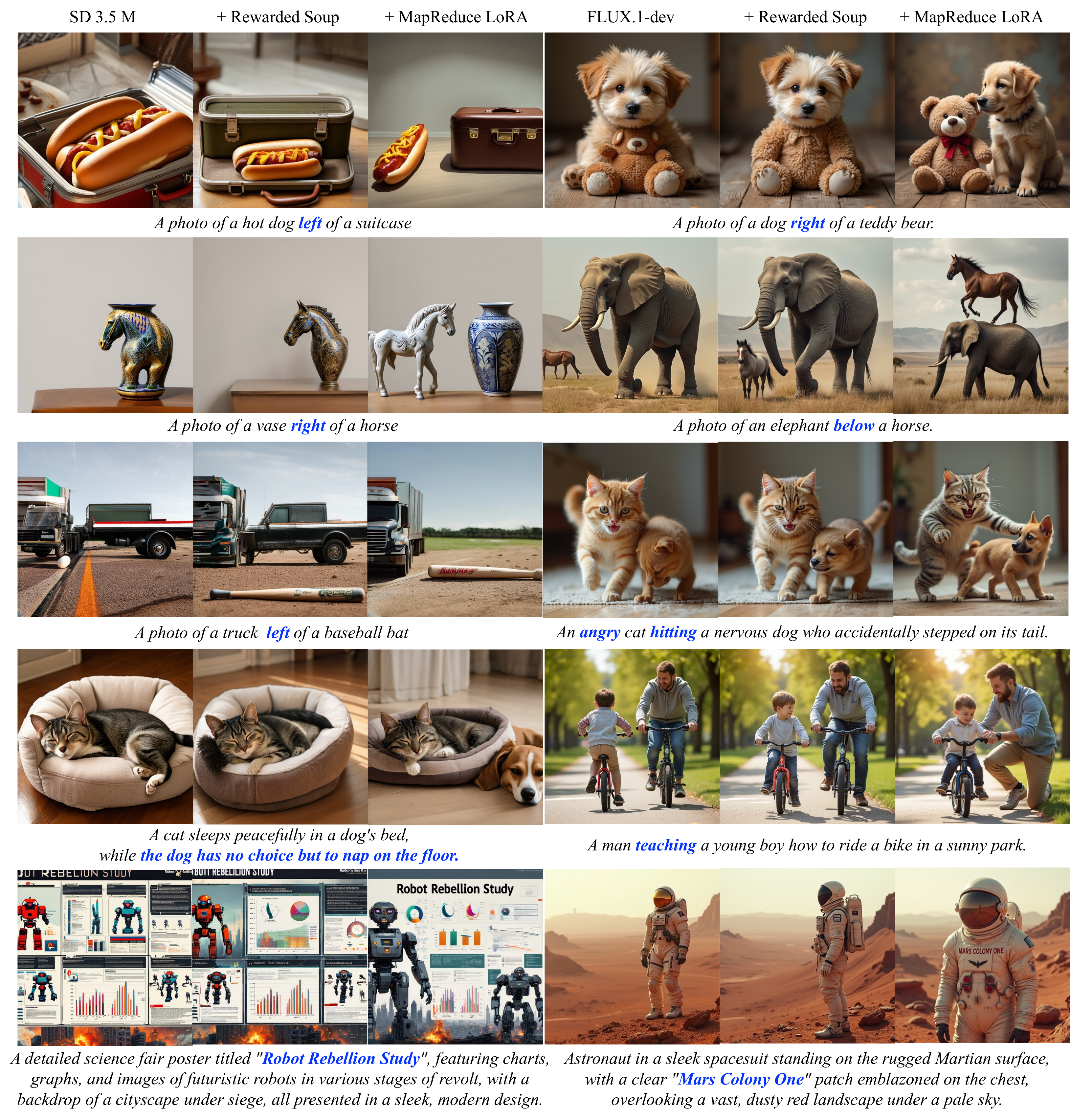}
    \vspace{-.7cm}
     \caption{\textbf{MapReduce LoRA} enhances the generative qualities, \ie, image aesthetics, positional relationship and text rendering quality, by optimizing the model with multiple rewards simultaneously.}  
     \label{fig:sd35_flux_qual}
     \vspace{-.3cm}
  \end{figure*}

\section{Experiments}
\label{sec:exp}

\begin{figure}[t]
    \centering
    \includegraphics[width=.95\linewidth]{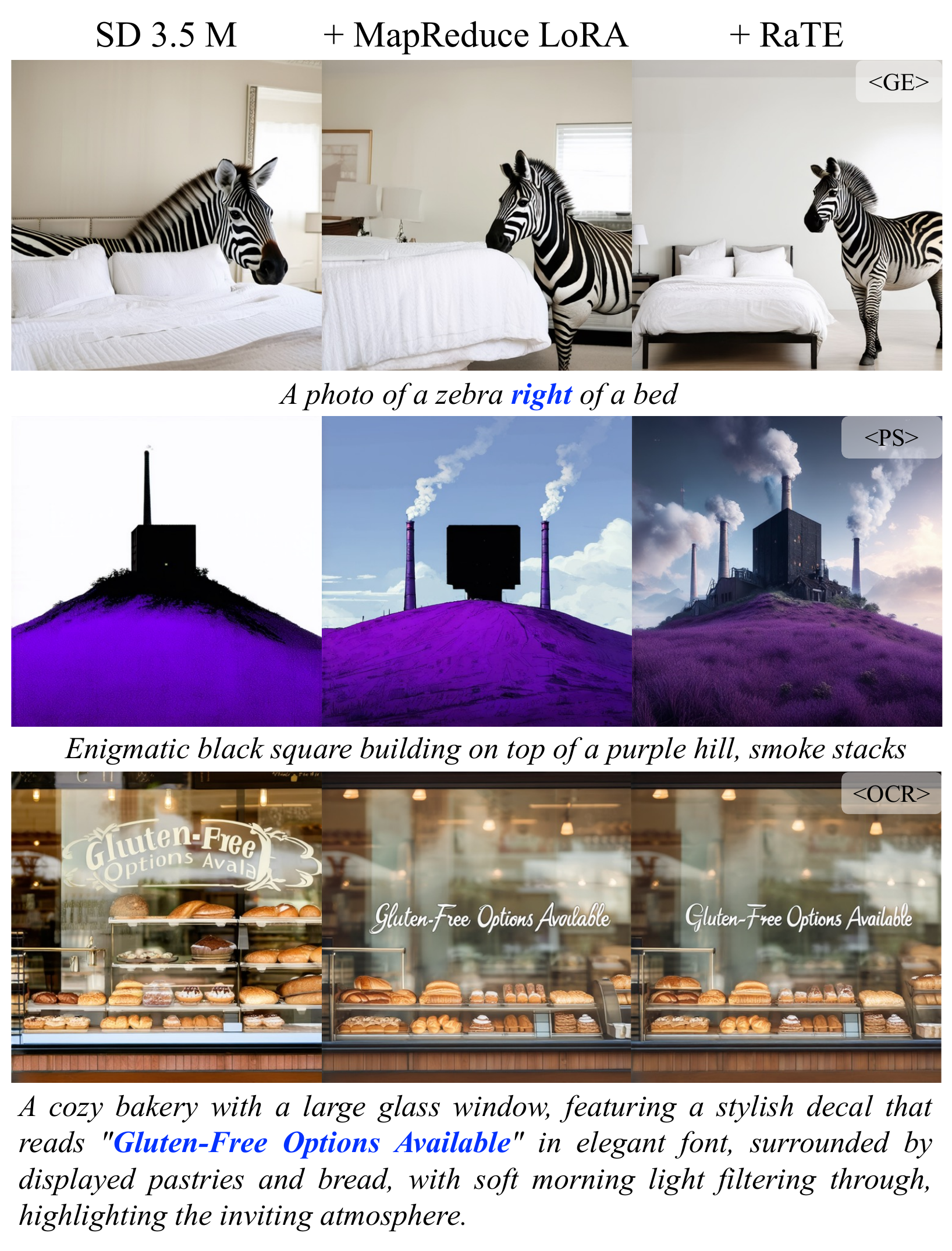}
    \vspace{-.3cm}
     \caption{\textbf{RaTE} enables per-reward control at inference.}  
     \label{fig:sd35m_rate}
     \vspace{-0.5cm}
  \end{figure}

\begin{table*}[t]
    \begin{center}
    \small
    \caption{Text-to-Image performance comparison on in-domain rewards (GenEval~\cite{GenEval_NeurIPS'23}, PickScore~\cite{PickScore_NeurIPS'23}, and OCR~\cite{paddleocr_arxiv'25}) within corresponding datasets and out-of-domain rewards (VQAScore~\cite{VQAScore_ECCV24}, MPS~\cite{MPS_CVPR'24}, and VILA~\cite{VILA_CVPR'23}) within PartiPrompts~\cite{partiprompt_TMLR22} and GenAI-Bench~\cite{GenAIBench_CVPRW24}. \textit{The performance is evaluated with fp32 precision. {\color{red}{Red color}} means the performance is degraded compared to the baseline.}}
    \label{tab:t2i_quan_id_ood}
    \vspace{-0.3cm}
    \scalebox{0.73}{%
    \setlength{\tabcolsep}{0.3em}
    \begin{tabular}{lcccccccc|c|c|ccc|c}
        \toprule
        \multirow{2}{*}{Method} & \multirow{2}{*}{Reward} & \multicolumn{7}{c}{GenEval~\cite{GenEval_NeurIPS'23}} & \multirow{2}{*}{PickScore} & \multirow{2}{*}{OCR} &  \multicolumn{3}{c}{PartiPrompts} & GenAIBench\\
        \cmidrule(lr){3-9}\cmidrule(lr){12-14}\cmidrule(lr){15-15}
        &  & Single Obj. & Two Obj. & Counting & Colors & Position & Color Attr. & Overall & & & VQAScore & MPS & VILA & VQAScore\\
        \midrule
        \rowcolor[HTML]{EFEFEF} 
        SD 3 M$^{*}$~\cite{sd3m_hf} & x  & 0.99 & 0.84 & 0.56 & 0.84 & 0.32 & 0.52 & 0.68 & - & - & 0.908 & 13.39 & 5.793 & - \\
        \multirow{3}{*}{SD 3 M + CaPO$^{*}$~\cite{CaPO_CVPR'25}} & VQAScore~\cite{VQAScore_ECCV24} & & & & & & & & & & & & \\
        & +MPS~\cite{MPS_CVPR'24} & 0.99 & 0.87 & 0.63 & 0.86 & \color{red}{0.31} & 0.59 & 0.71 & - & - & 0.914 & 13.58 & 5.943 & -\\
        & +VILA~\cite{VILA_CVPR'23} & \scriptsize{\textcolor{Gray}{(+0.00\%)}}	& \scriptsize{\textcolor{blue}{(+3.57\%)}}	& \scriptsize{\textcolor{blue}{(+12.50\%)}}	& \scriptsize{\textcolor{blue}{(+2.38\%)}}	& \scriptsize{\textcolor{red}{(-3.13\%)}}	& \scriptsize{\textcolor{blue}{(+13.46\%)}}	& \scriptsize{\textcolor{blue}{(+4.41\%)}}	& -	& -	& \scriptsize{\textcolor{blue}{(+0.66\%)}}	& \scriptsize{\textcolor{blue}{(+1.42\%)}}	& \scriptsize{\textcolor{blue}{(+2.59\%)}} & - \\
        \midrule
        \rowcolor[HTML]{EFEFEF} 
        SD 3.5 M~\cite{sd35m_hf} & x & 1.00	& 0.86	& 0.55	& 0.82	& 0.23	& 0.62	& 0.68	& 21.784	& 0.601 & 0.861 & 11.68 & 6.034 & 0.744\\
        \multirow{3}{*}{\makecell{Individual Experts\\(Flow-GRPO~\cite{flow_grpo_arxiv'25})}} & GenEval~\cite{GenEval_NeurIPS'23} & 1.00 & 0.96	& 0.93	& 0.91	& 0.93	& 0.84	& 0.93	& 21.852	& 0.654 & 0.879	& 11.81& 6.101 & 0.782 \\
        & PickScore~\cite{PickScore_NeurIPS'23}  & 1.00 &	0.94	& 0.76	& 0.88	& 0.36	& \color{red}{0.59}	& 0.76	& 23.359	& 0.716 & 0.880	& 12.66	& 6.755 & 0.774 \\
        & OCR~\cite{paddleocr_arxiv'25}  & 1.00 &	0.87	& 0.56	& 0.82	& 0.25	& \color{red}{0.59}	& 0.68	& 21.800	& 0.934 & 0.863 & 11.75 & \color{red}{6.014} & 0.750\\
        Rewarded Soup~\cite{reward_soup_nips'23} & \cite{GenEval_NeurIPS'23} + \cite{PickScore_NeurIPS'23} + \cite{paddleocr_arxiv'25}  & 1.00	& 0.94	& 0.76	& 0.86	& 0.43	& 0.72	& 0.79	& 22.276	& 0.805 & 0.875	& 12.13	& 6.194 & 0.772\\
        \midrule
        MORL-D$^{\alpha}$ & \cite{GenEval_NeurIPS'23} + \cite{PickScore_NeurIPS'23} + \cite{paddleocr_arxiv'25}  & 1.00 & 0.99 & 0.94 & 0.89 & 0.88 & 0.78 & 0.91 & 22.067 & 0.942 & 0.881 & 12.00 & 6.113 &	0.777\\
        MORL-DR$^{\alpha}$ & \cite{GenEval_NeurIPS'23} + \cite{PickScore_NeurIPS'23} + \cite{paddleocr_arxiv'25}  & 1.00 & \color{red}{0.82} & 0.61 & \color{red}{0.81} & \color{red}{0.21} & \color{red}{0.59} & \color{red}{0.67} & 21.834 & 0.641 & \color{red}{0.860}& 11.71 & \color{red}{6.014} & 0.746\\
        \lgcell{MapReduce LoRA} & \lgcell{\cite{GenEval_NeurIPS'23} + \cite{PickScore_NeurIPS'23} + \cite{paddleocr_arxiv'25}}  & \lgcell{1.00}	& \lgcell{0.98}	& \lgcell{0.93}	& \lgcell{0.88}	& \lgcell{0.81}	& \lgcell{0.78}	& \lgcell{0.90}	& \lgcell{22.552}	& \lgcell{0.923} & \lgcell{0.885} & \lgcell{12.25} & \lgcell{6.266} & \lgcell{0.782}\\
        &  & \scriptsize{\textcolor{Gray}{(+0.00\%)}}	& \scriptsize{\textcolor{blue}{(+14.12\%)}}	& \scriptsize{\textcolor{blue}{(+68.18\%)}}	& \scriptsize{\textcolor{blue}{(+7.80\%)}}	& \scriptsize{\textcolor{blue}{(+252.17\%)}}	& \scriptsize{\textcolor{blue}{(+25.81\%)}}	& \scriptsize{\textcolor{blue}{(+31.88\%)}}	& \scriptsize{\textcolor{blue}{(+3.53\%)}}	& \scriptsize{\textcolor{blue}{(+53.55\%)}}	& \scriptsize{\textcolor{blue}{(+2.78\%)}}	& \scriptsize{\textcolor{blue}{(+4.89\%)}}	& \scriptsize{\textcolor{blue}{(+3.85\%)}} & \scriptsize{\textcolor{blue}{(+5.13\%)}}\\
        \lgcell{MapReduce LoRA + RaTE} & \lgcell{\cite{GenEval_NeurIPS'23} + \cite{PickScore_NeurIPS'23} + \cite{paddleocr_arxiv'25}}  & \lgcell{1.00}	& \lgcell{0.98}	& \lgcell{0.93}	& \lgcell{0.90}	& \lgcell{0.95}	& \lgcell{0.79}	& \lgcell{0.92}	& \lgcell{22.777}	& \lgcell{0.936} & \lgcell{0.877} & \lgcell{12.44} & \lgcell{7.238} & \lgcell{0.777} \\ 
        & & \scriptsize{\textcolor{Gray}{(+0.00\%)}}	& \scriptsize{\textcolor{blue}{(+14.12\%)}}	& \scriptsize{\textcolor{blue}{(+68.18\%)}}	& \scriptsize{\textcolor{blue}{(+10.40\%)}}	& \scriptsize{\textcolor{blue}{(+313.04\%)}}	& \scriptsize{\textcolor{blue}{(+27.42\%)}}	& \scriptsize{\textcolor{blue}{(+36.08\%)}}	& \scriptsize{\textcolor{blue}{(+4.56\%)}}	& \scriptsize{\textcolor{blue}{(+55.72\%)}}	& \scriptsize{\textcolor{blue}{(+1.85\%)}} & \scriptsize{\textcolor{blue}{(+6.49\%)}} & \scriptsize{\textcolor{blue}{(+19.96\%)}} & \scriptsize{\textcolor{blue}{(+4.37\%)}} \\ 
        \midrule
        \rowcolor[HTML]{EFEFEF} 
        FLUX.1-dev~\cite{flux24} & x  & 0.99 & 0.87 & 0.71 & 0.82 & 0.18 & 0.44 & 0.67 & 22.006 & 0.573 & 0.830 & 12.37 & 6.629 &	0.737\\
        \multirow{3}{*}{\makecell{Individual Experts\\(Flow-GRPO~\cite{flow_grpo_arxiv'25})}} & GenEval~\cite{GenEval_NeurIPS'23}  & 1.00 &	1.00 & 0.90 & 0.89 & 0.90 & 0.73 & 0.90 & 22.057 & 0.692 & 0.854	& 12.45 & 6.661 &	0.764\\
        & PickScore~\cite{PickScore_NeurIPS'23}  & 1.00 & 0.95	& \color{red}{0.61}	& 0.85	& 0.24	& 0.57	& 0.70	& 23.566	& 0.723 & 0.875	& 12.76	& \color{red}{5.514} &	0.772\\
        & OCR~\cite{paddleocr_arxiv'25} & 1.00	& \color{red}{0.84}	& 0.76	& 0.83	& 0.20	& 0.46	& 0.68	& \color{red}{21.997}	& 0.971 & 0.841 &  \color{red}{12.32} & \color{red}{6.591} &	0.749 \\
        Rewarded Soup~\cite{reward_soup_nips'23} & \cite{GenEval_NeurIPS'23} + \cite{PickScore_NeurIPS'23} + \cite{paddleocr_arxiv'25}  & 0.99 & 0.96 & 0.81	& 0.88	& 0.31	& 0.66	& 0.77	& 22.503	& 0.868 & 0.862 & 12.73 & 6.650 &	0.767\\
        \midrule
        MORL-D$^{\alpha}$ & \cite{GenEval_NeurIPS'23} + \cite{PickScore_NeurIPS'23} + \cite{paddleocr_arxiv'25}  & 1.00 & 	0.99 &	0.89 &	0.90 &	0.78 &	0.73 &	0.88 &	22.026 &	0.932 & 0.834 &	\color{red}{12.36} &	\color{red}{6.513} &	\color{red}{0.733}\\
        MORL-DR$^{\alpha}$ & \cite{GenEval_NeurIPS'23} + \cite{PickScore_NeurIPS'23} + \cite{paddleocr_arxiv'25}  & 1.00 &	0.95 &	0.90 &	0.94 &	0.73 &	0.79 &	0.88 &	22.112 &	0.952 & 0.840 &	12.41 &	\color{red}{6.534} &	0.745 \\
        \lgcell{MapReduce LoRA} & \lgcell{\cite{GenEval_NeurIPS'23} + \cite{PickScore_NeurIPS'23} + \cite{paddleocr_arxiv'25}}  & \lgcell{1.00}	& \lgcell{0.96}	& \lgcell{0.88}	& \lgcell{0.88}	& \lgcell{0.81}	& \lgcell{0.79}	& \lgcell{0.89}	& \lgcell{22.951} & \lgcell{0.957} & \lgcell{0.875} & \lgcell{12.95}	& \lgcell{\color{red}{6.572}} &	\lgcell{0.775}\\
        & 	 & \scriptsize{\textcolor{blue}{(+1.27\%)}}	& \scriptsize{\textcolor{blue}{(+10.46\%)}}	& \scriptsize{\textcolor{blue}{(+22.81\%)}}	& \scriptsize{\textcolor{blue}{(+7.80\%)}}	& \scriptsize{\textcolor{blue}{(+350.00\%)}}	& \scriptsize{\textcolor{blue}{(+79.55\%)}}	& \scriptsize{\textcolor{blue}{(+32.68\%)}}	& \scriptsize{\textcolor{blue}{(+4.29\%)}}	& \scriptsize{\textcolor{blue}{(+67.06\%)}}	& \scriptsize{\textcolor{blue}{(+5.49\%)}}	 & \scriptsize{\textcolor{blue}{(+4.70\%)}}	 & \scriptsize{\textcolor{red}{(-0.87\%)}} & \scriptsize{\textcolor{blue}{(+5.09\%)}} \\
        \bottomrule
    \end{tabular}}
    \end{center}
    \vspace{-0.3cm}
    \footnotesize{$^{*}$ From CaPO Table 2 and 3 \cite{CaPO_CVPR'25}; as the code is not released, we report their numbers and cannot compare CaPO on SD 3.5 M~\cite{sd35m_hf} and FLUX.1-dev~\cite{flux24}. \\$^{\alpha}$ The detailed settings of MORL-D/DR: Please refer to the Supplement~\ref{supp:sec:implementation_details}.}\\
\end{table*}

\begin{figure*}[t]
    \centering
    \vspace{0.3em}
    \begin{minipage}[b]{0.02\textwidth}~\end{minipage}
    \begin{minipage}[b]{0.32\textwidth}\centering \small GenEval\end{minipage}\hfill
    \begin{minipage}[b]{0.32\textwidth}\centering \small PickScore\end{minipage}\hfill
    \begin{minipage}[b]{0.32\textwidth}\centering \small OCR\end{minipage}
    
    \vspace{.3em}
    \begin{minipage}[c]{0.02\textwidth}
        \centering
        \rotatebox{90}{\small \qquad k = 1}
    \end{minipage}
    \begin{minipage}[c]{0.32\textwidth}
        \centering
        \includegraphics[width=\linewidth]{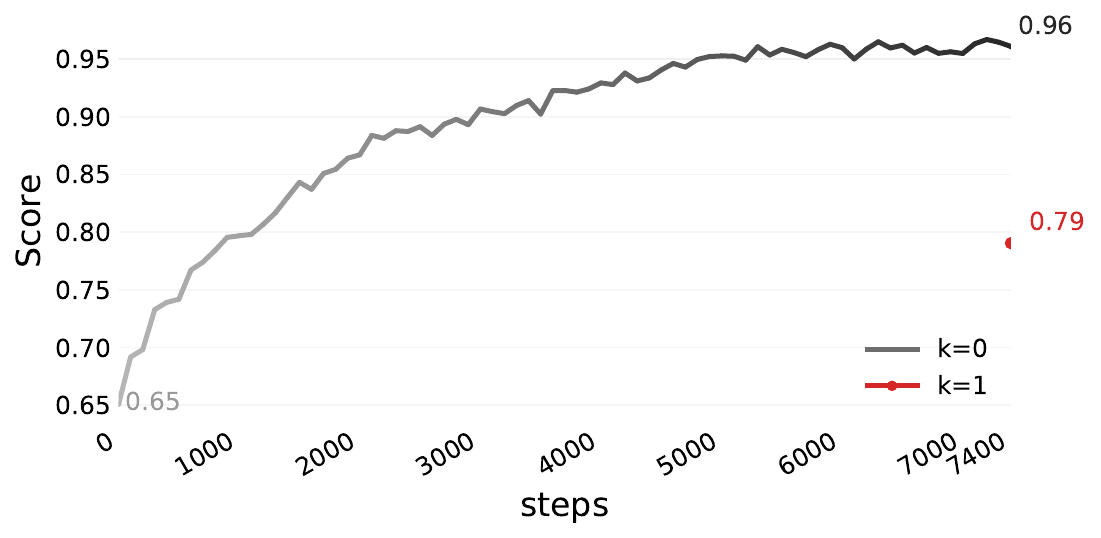}
    \end{minipage}\hfill
    \begin{minipage}[c]{0.32\textwidth}
        \centering
        \includegraphics[width=\linewidth]{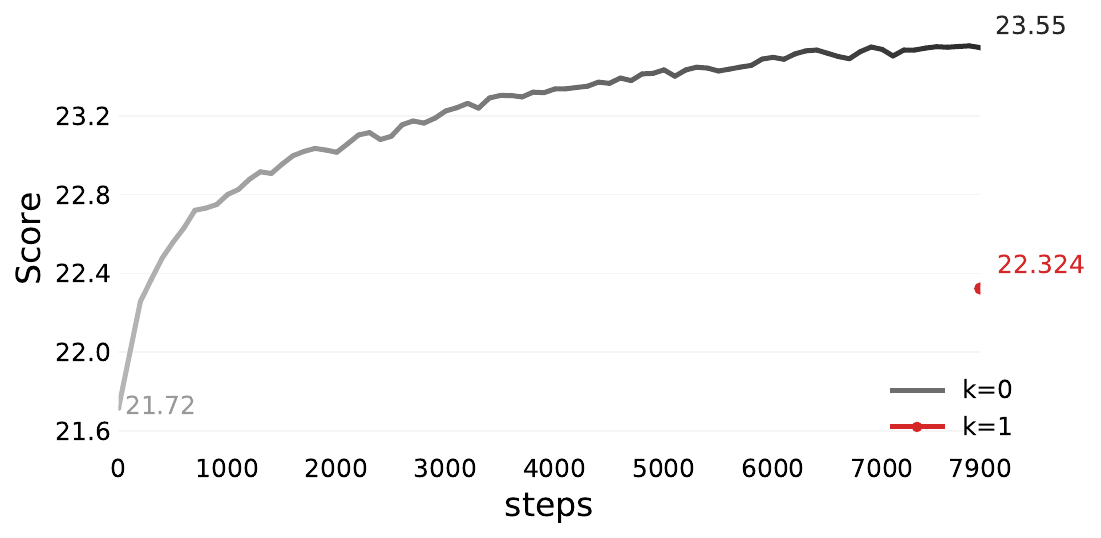}
    \end{minipage}\hfill
    \begin{minipage}[c]{0.32\textwidth}
        \centering
        \includegraphics[width=\linewidth]{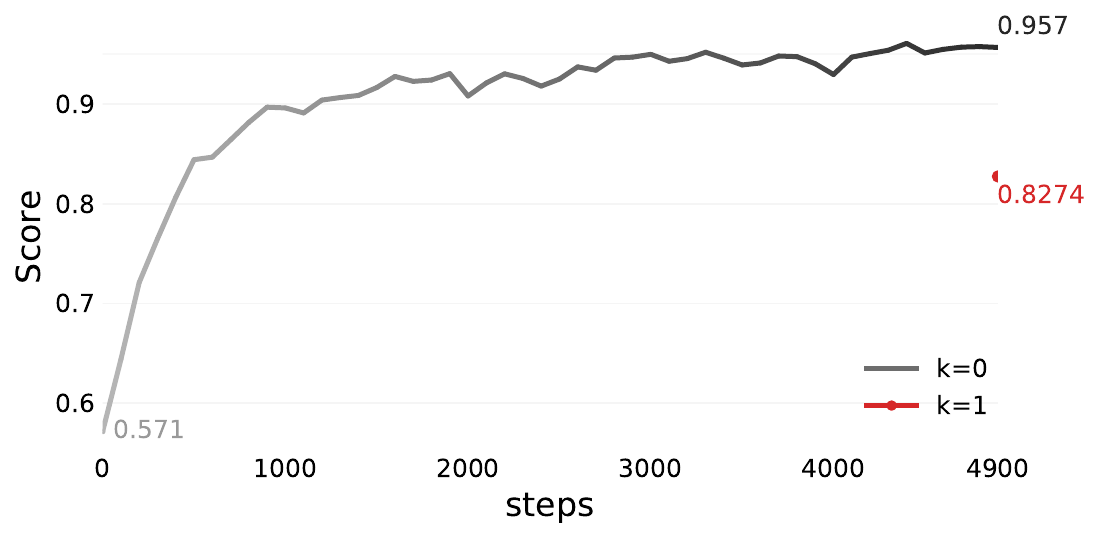}
    \end{minipage}

    \begin{minipage}[c]{0.02\textwidth}
        \centering
        \rotatebox{90}{\small \qquad k = 4}
    \end{minipage}
    \begin{minipage}[c]{0.32\textwidth}
        \centering
        \includegraphics[width=\linewidth]{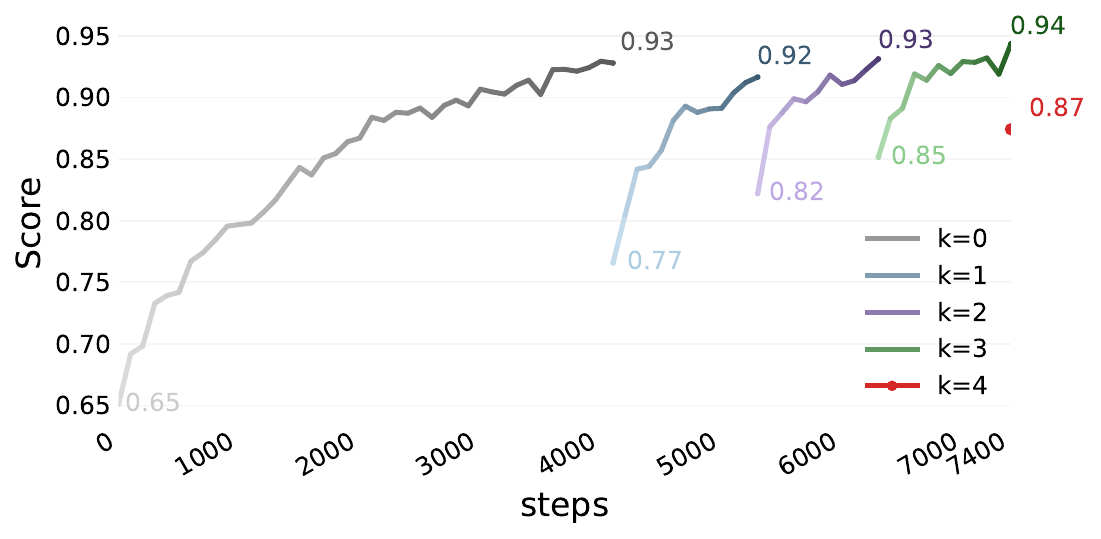}
    \end{minipage}\hfill
    \begin{minipage}[c]{0.32\textwidth}
        \centering
        \includegraphics[width=\linewidth]{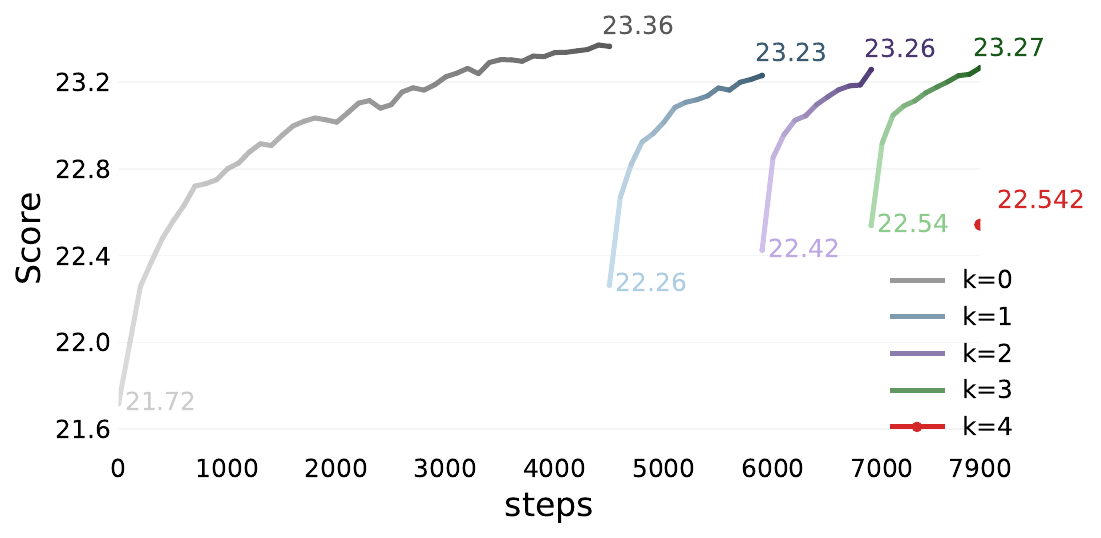}
    \end{minipage}\hfill
    \begin{minipage}[c]{0.32\textwidth}
        \centering
        \includegraphics[width=\linewidth]{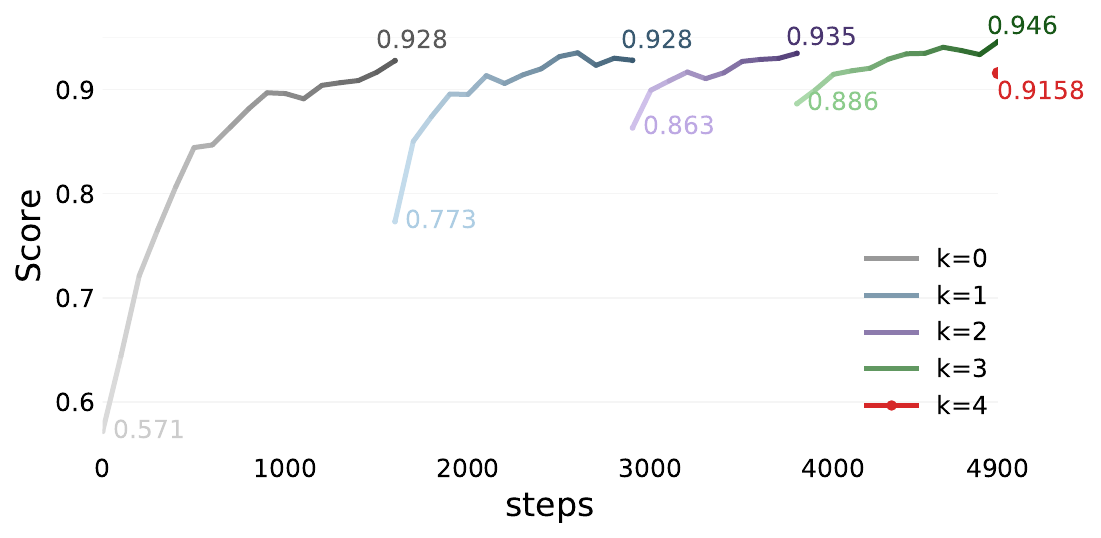}
    \end{minipage}

    \begin{minipage}[c]{0.02\textwidth}
        \centering
        \rotatebox{90}{\small \qquad k = 10}
    \end{minipage}
    \begin{minipage}[c]{0.32\textwidth}
        \centering
        \includegraphics[width=\linewidth]{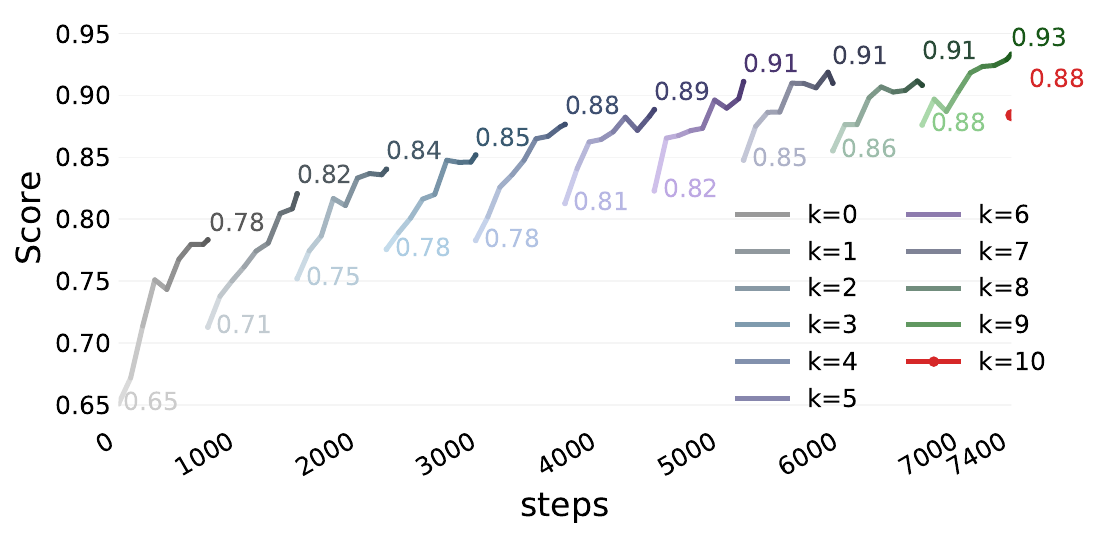}
    \end{minipage}\hfill
    \begin{minipage}[c]{0.32\textwidth}
        \centering
        \includegraphics[width=\linewidth]{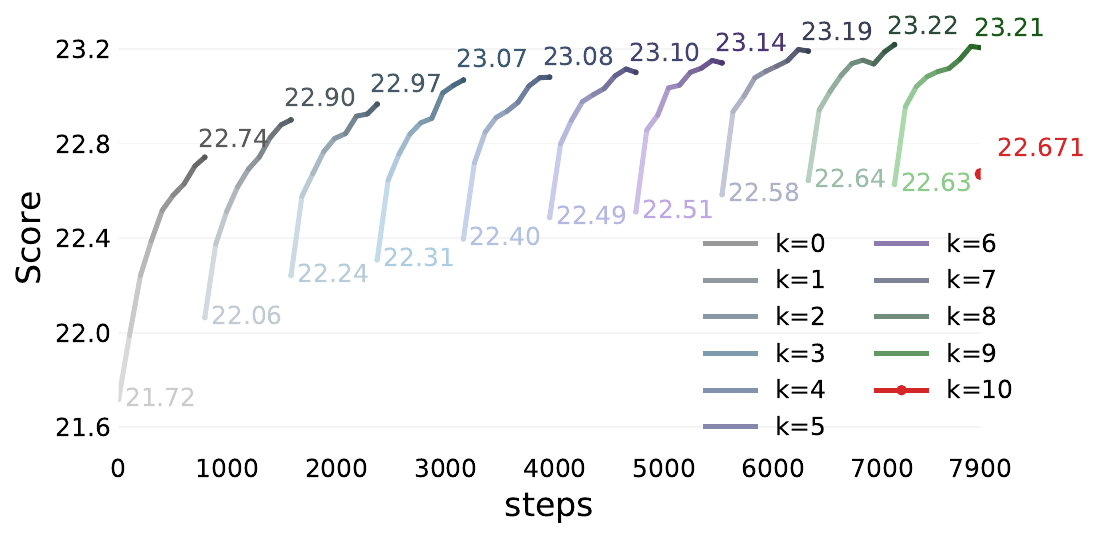}
    \end{minipage}\hfill
    \begin{minipage}[c]{0.32\textwidth}
        \centering
        \includegraphics[width=\linewidth]{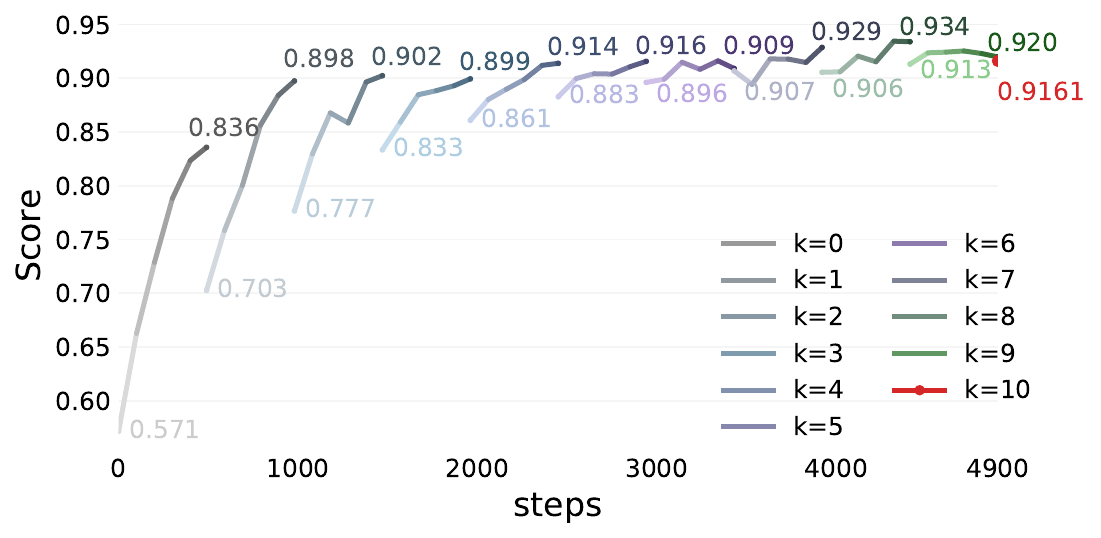}
    \end{minipage}
    \vspace{-.3cm}
    \caption{
        \textbf{Ablation study on merging iterations} ($k = 1$ vs. $4$ vs. $10$).
        \textit{The performance is evaluated during training with mixed precision, 
        where the text encoder is set to fp16 precision and the rest of the model to fp32.}
    }
    \vspace{-0.5cm}
    \label{fig:exp:ablation_merging_10}
\end{figure*}

\subsection{Experimental Setup}

\noindent\textbf{Models and datasets.} For \textit{Text-to-Image generation}, we use Stable Diffusion 3.5 Medium (SD 3.5 M)~\cite{sd35m_hf} and FLUX.1-dev~\cite{flux24} as base models. We adopt the same datasets as Flow-GRPO~\cite{flow_grpo_arxiv'25}: GenEval~\cite{GenEval_NeurIPS'23}, PickScore~\cite{PickScore_NeurIPS'23}, and OCR~\cite{paddleocr_arxiv'25}. We compare against CaPO~\cite{CaPO_CVPR'25}, Flow-GRPO individual experts~\cite{flow_grpo_arxiv'25}, multi-objective RL with mixed data (MORL-D) and with mixed data plus mixed rewards (MORL-DR), and Rewarded Soup~\cite{reward_soup_nips'23}. The merging iteration is 4 for all reported results if not specified.
For \textit{Text-to-Video generation}, we use the HunyuanVideo backbone~\cite{hunyuanvideo_arxiv'25}. We follow the Dance-GRPO~\cite{dancegrpo_arxiv25} GitHub recipe (optimizer, GRPO grouping, scheduler, data sampling), but apply parameter-efficient LoRA instead of full fine-tuning. We compare to Dance-GRPO individual experts~\cite{dancegrpo_arxiv25}. Because the GitHub configuration differs from the paper, we report reproduced baselines trained on the released data for fairness. The merging iteration is 3 for all reported results if not specified.
For \textit{Language tasks}, we use llama2-7B~\cite{llama2_arxiv23} as base model, following Bone Soup~\cite{bonesoup_acl25} setup. We first train the Llama2-7B~\cite{llama2_arxiv23} backbone via supervised fine-tuning (SFT) using the same datasets and hyperparameters as Bone Soup~\cite{bonesoup_acl25}, then train LoRA experts with PPO~\cite{ppo_arxiv17} under the same settings of datasets and training hyperparameters. The merging iteration is 3 for all reported results if not specified. 
\textit{For implementation details, please refer to the Supplement~\ref{supp:sec:implementation_details}.}

\noindent\textbf{Reward models.} For \textit{Text-to-Image}, we train with three rewards: detection-based text-image alignment (GenEval~\cite{GenEval_NeurIPS'23}), human preference (PickScore~\cite{PickScore_NeurIPS'23}), and text rendering quality (OCR~\cite{paddleocr_arxiv'25}). For evaluation, we additionally report clip-based text-image alignment (VQAScore~\cite{VQAScore_ECCV24}), human preference (MPS~\cite{MPS_CVPR'24}), and aesthetic quality (VILA~\cite{VILA_CVPR'23}). For \textit{Text-to-Video}, we use two rewards: Visual Quality (VQ) and Motion Quality (MQ) from VideoAlign~\cite{videoalign_arxiv25}. 
For the \textit{Language task}, we employ four rewards across two tasks: Reddit Summary (Faithful and Preference1) and Helpful Assistant (Helpful and Harmless).

\subsection{Qualitative Results}

Fig.~\ref{fig:sd35m_merge_progress} shows iterative improvements from MapReduce LoRA and the controllability of RaTE (see also Fig.~\ref{fig:sd35m_rate}). Compared to the base model and Rewarded Soup~\cite{reward_soup_nips'23} (Fig.~\ref{fig:sd35_flux_qual}), MapReduce LoRA better handles challenging cases, including uncommon spatial relations (\eg, an elephant below a horse) and high-quality rendering of small text. \textit{For more qualitative results on Text-to-Image and Text-to-Video, please refer to the Supplement~\ref{supp:t2i_more_qual} and \ref{supp:t2v_more_qual}.}

\subsection{Quantitative Results}
\noindent\textbf{MapReduce LoRA.} Table~\ref{tab:t2i_quan_id_ood} reports \textit{Text-to-Image} performance on GenEval~\cite{GenEval_NeurIPS'23}, PickScore~\cite{PickScore_NeurIPS'23}, OCR~\cite{paddleocr_arxiv'25}, VQAScore~\cite{VQAScore_ECCV24}, MPS~\cite{MPS_CVPR'24}, and VILA~\cite{VILA_CVPR'23} using SD 3.5 M~\cite{sd35m_hf} and FLUX.1-dev~\cite{flux24}.
MapReduce LoRA improves both in-domain and out-of-domain metrics. Relative to CaPO~\cite{CaPO_CVPR'25}, it delivers larger gains across the four CaPO metrics. On CaPO's in-domain metrics (VQAScore~\cite{VQAScore_ECCV24}, MPS~\cite{MPS_CVPR'24}, VILA~\cite{VILA_CVPR'23}), which are out-of-domain for us, our improvements are 2.78\%, 4.89\%, and 3.85\%, exceeding CaPO's 0.66\%, 1.42\%, and 2.59\%. On CaPO's out-of-domain metric (GenEval~\cite{GenEval_NeurIPS'23}), our improvement is 31.88\% versus CaPO's 4.41\%. Flow-GRPO individual experts~\cite{flow_grpo_arxiv'25}, each tuned to a single reward, substantially boost the targeted metric but transfer poorly to others. MORL-D and MORL-DR can be unstable under conflicting objectives; for instance, PickScore often collapses during joint training and is dominated by other rewards (\eg, OCR). In contrast, MapReduce LoRA maintains competitive performance across all targeted rewards.

Table~\ref{tab:t2v_mq_vq} reports \textit{Text-to-Video} results. We evaluate 1,024 held-out prompts (fixed once). For each prompt, we generate four samples using a fixed set of seeds shared across models and report mean Visual Quality (VQ) and Motion Quality (MQ) from VideoAlign~\cite{videoalign_arxiv25}. MapReduce LoRA improves both VQ and MQ (+48.09\% and +89.96\%) over individually trained experts (+34.55\% and +74.10\%).

In Figs.~\ref{supp:fig:reddit_summary_pf_llm} and \ref{supp:fig:assistant_pf_llm}, we compare MapReduce LoRA against Llama2 (base)~\cite{llama2_arxiv23}, Llama2 after SFT, Rewarded Soup~\cite{reward_soup_nips'23}, and Bone Soup~\cite{bonesoup_acl25}. MapReduce LoRA achieves state-of-the-art performance on both tasks, underscoring its cross-modal generality.

\begin{table}[h]
    \begin{center}
    \caption{Text-to-Video comparison on Visual and Motion Quality.}
    \label{tab:t2v_mq_vq}
    \vspace{-0.2cm}
    \small
    \scalebox{.85}{%
    \setlength{\tabcolsep}{0.3em}
    \begin{tabular}{lccc}
        \toprule
        Method & Reward & VQ & MQ\\
        \midrule
        \rowcolor[HTML]{EFEFEF} 
        HunyuanVideo~\cite{hunyuanvideo_arxiv'25} & x & 3.25 & 0.95 \\
        \multirow{2}{*}{\makecell{Individual Experts\\(DanceGRPO~\cite{dancegrpo_arxiv25})}} & VQ & 4.37 \scriptsize{\textcolor{blue}{(+34.55\%)}} & 1.20 \scriptsize{\textcolor{blue}{(+26.41\%)}}\\
         & MQ & 3.80 \scriptsize{\textcolor{blue}{(+16.95\%)}}& 1.66 \scriptsize{\textcolor{blue}{(+74.10\%)}}\\
        Rewarded Soup~\cite{reward_soup_nips'23} & VQ+MQ & 4.13 \scriptsize{\textcolor{blue}{(+27.37\%)}} & 1.43 \scriptsize{\textcolor{blue}{(+50.42\%)}} \\
        MapReduce LoRA  & VQ+MQ & 4.81 \scriptsize{\textcolor{blue}{(+48.09\%)}} & 1.81 \scriptsize{\textcolor{blue}{(+89.96\%)}} \\
        \bottomrule
    \end{tabular}}
    \end{center}
    \vspace{-0.5cm}
\end{table}

\noindent\textbf{[Ablation] MapReduce LoRA: Effect of merging iterations under a fixed training budget.} With the total training steps fixed (Fig.~\ref{fig:exp:ablation_merging_10}), we compare merging iterations \(k \in \{1,4,10\}\). Increasing \(k\) consistently improves final performance and reduces degradation from merging multiple experts. 
For \(k=10\) relative to \(k=4\) and \(k=1\), the gains are +1.12\% and +11.84\% on GenEval~\cite{GenEval_NeurIPS'23}, +0.57\% and +1.56\% on PickScore~\cite{PickScore_NeurIPS'23}, and +0.03\% and +10.72\% on OCR~\cite{paddleocr_arxiv'25}.

\noindent\textbf{Reward-aware token embedding (RaTE).} 
Tables~\ref{tab:t2i_quan_id_ood} and \ref{tab:rate_performance} report RaTE performance on GenEval~\cite{GenEval_NeurIPS'23}, PickScore~\cite{PickScore_NeurIPS'23}, and OCR~\cite{paddleocr_arxiv'25} using SD 3.5 M~\cite{sd35m_hf}. At inference, RaTE appends a trained control token to the end of the prompt based on the specified preference. RaTE enables per-reward control and jointly improves multiple rewards. In Table~\ref{tab:t2i_quan_id_ood}, token gains are orthogonal to transformer tuning: RaTE adds 4.20\%, 1.03\%, and 2.17\% on GenEval~\cite{GenEval_NeurIPS'23}, PickScore~\cite{PickScore_NeurIPS'23}, and OCR~\cite{paddleocr_arxiv'25}, respectively. In Table~\ref{tab:rate_performance}, when using all three tokens, the first appended token accounts for most of the gain. 
RaTE enables lightweight, composable inference-time control and costs only 0.1579× of MapReduce LoRA to train.
Token control is effective on Stable Diffusion~\cite{sd35m_hf, ti_iclr'23, CAT_NeurIPS24}, which uses explicit cross-attention for text conditioning. In contrast, FLUX.1-dev~\cite{flux24} performs joint text–image sequence modeling, causing a modified token embedding to perturb both text and image tokens across layers, which makes reward-specific token control substantially less stable. We leave this for future exploration.

\begin{table}[h]
    \begin{center}
    \caption{\textbf{Reward-aware Token Embedding results.} \small{$<$}\texttt{GE}\small{$>$}, \small{$<$}\texttt{PS}\small{$>$}, and \small{$<$}\texttt{OCR}\small{$>$} denote tokens trained on GenEval~\cite{GenEval_NeurIPS'23}, PickScore~\cite{PickScore_NeurIPS'23}, and OCR~\cite{paddleocr_arxiv'25}, respectively. {\color{gray!50}{Gray}} indicates not trained on that reward; \textbf{bold} indicates the best among variants.}
    \label{tab:rate_performance}
    \vspace{-0.2cm}
    \small
    \scalebox{.85}{%
    \setlength{\tabcolsep}{0.3em}
    \begin{tabular}{lccc}
        \toprule
        Method & GenEval~\cite{GenEval_NeurIPS'23} & PickScore~\cite{PickScore_NeurIPS'23} & OCR~\cite{paddleocr_arxiv'25}\\
        \midrule
        \rowcolor[HTML]{EFEFEF} 
        SD 3.5 M~\cite{sd35m_hf}  & 0.68 & 21.784 & 0.601 \\
        + \small{$<$}\texttt{GE}\small{$>$}            & 0.76 & \color{gray!50}{21.732} & \color{gray!50}{0.601} \\
        + \small{$<$}\texttt{PS}\small{$>$}             & \color{gray!50}{0.69} & 22.052 & \color{gray!50}{0.591} \\
        + \small{$<$}\texttt{OCR}\small{$>$}            & \color{gray!50}{0.68} & \color{gray!50}{21.724} & 0.623 \\
        \midrule
        + \small{$<$}\texttt{GE}\small{$>$} + \small{$<$}\texttt{PS}\small{$>$}   & \textbf{0.77} & 21.990 & \color{gray!50}{0.610} \\
        + \small{$<$}\texttt{GE}\small{$>$} + \small{$<$}\texttt{OCR}\small{$>$}  & \textbf{0.77} & \color{gray!50}{21.690} & \textbf{0.633} \\
        + \small{$<$}\texttt{PS}\small{$>$} + \small{$<$}\texttt{GE}\small{$>$}   & 0.74 & 22.014 & \color{gray!50}{0.599} \\
        + \small{$<$}\texttt{PS}\small{$>$} + \small{$<$}\texttt{OCR}\small{$>$}  & \color{gray!50}{0.69} & \textbf{22.064} & 0.592 \\
        + \small{$<$}\texttt{OCR}\small{$>$} + \small{$<$}\texttt{GE}\small{$>$}  & 0.75 & \color{gray!50}{21.736} & 0.619 \\
        + \small{$<$}\texttt{OCR}\small{$>$} + \small{$<$}\texttt{PS}\small{$>$}  & \color{gray!50}{0.69} & 22.040 & 0.607 \\
        \midrule
        + \small{$<$}\texttt{GE}\small{$>$} + \small{$<$}\texttt{PS}\small{$>$} + \small{$<$}\texttt{OCR}\small{$>$} & 0.76 & 21.992 & 0.613 \\
        + \small{$<$}\texttt{GE}\small{$>$} + \small{$<$}\texttt{OCR}\small{$>$} + \small{$<$}\texttt{PS}\small{$>$} & 0.75 & 21.981 & 0.615 \\
        + \small{$<$}\texttt{PS}\small{$>$} + \small{$<$}\texttt{GE}\small{$>$} + \small{$<$}\texttt{OCR}\small{$>$} & 0.75 & 22.009 & 0.606 \\
        + \small{$<$}\texttt{PS}\small{$>$} + \small{$<$}\texttt{OCR}\small{$>$} + \small{$<$}\texttt{GE}\small{$>$} & 0.72 & 22.027 & 0.606 \\
        + \small{$<$}\texttt{OCR}\small{$>$} + \small{$<$}\texttt{GE}\small{$>$} + \small{$<$}\texttt{PS}\small{$>$} & 0.73 & 22.013 & 0.617 \\
        + \small{$<$}\texttt{OCR}\small{$>$} + \small{$<$}\texttt{PS}\small{$>$} + \small{$<$}\texttt{GE}\small{$>$} & 0.74 & 22.014 & 0.614 \\
        \bottomrule
    \end{tabular}}
    \end{center}
    \vspace{-0.6cm}
\end{table}



\vspace{.3cm}

\noindent\textbf{[Ablation] RaTE: Effect of appended token count.} Table~\ref{tab:ablation_rate_num} reports RaTE performance as we vary the number of appended tokens. In the table, \small{$<$}\texttt{RaTE}\small{$>$} denotes the metric-specific token: \small{$<$}\texttt{GE}\small{$>$} for GenEval~\cite{GenEval_NeurIPS'23}, \small{$<$}\texttt{PS}\small{$>$} for PickScore~\cite{PickScore_NeurIPS'23}, and \small{$<$}\texttt{OCR}\small{$>$} for OCR~\cite{paddleocr_arxiv'25}. Performance peaks differ by reward: GenEval saturates at 2--3 tokens (0.78), PickScore peaks at 1 (22.052), and OCR peaks at 3 (0.635).

\begin{table}[h]
    \begin{center}
    \caption{Ablation on the number of appended RaTE tokens.}
    \label{tab:ablation_rate_num}
    \vspace{-.2cm}
    \scalebox{.85}{%
    \setlength{\tabcolsep}{0.3em}
    \begin{tabular}{lccc}
        \toprule
        Method & GenEval~\cite{GenEval_NeurIPS'23} & PickScore~\cite{PickScore_NeurIPS'23} & OCR~\cite{paddleocr_arxiv'25}\\
        \midrule
        \rowcolor[HTML]{EFEFEF} 
        SD 3.5 M~\cite{sd35m_hf}  & 0.68 & 21.784 & 0.601 \\
        + \small{$<$}\texttt{RaTE}\small{$>$} x1   & 0.76 & \textbf{22.052} & 0.623 \\
        + \small{$<$}\texttt{RaTE}\small{$>$} x2   & \textbf{0.78} & 22.037 & 0.621 \\
        + \small{$<$}\texttt{RaTE}\small{$>$} x3   & \textbf{0.78} & 22.021 & \textbf{0.635} \\
        + \small{$<$}\texttt{RaTE}\small{$>$} x10  & \textbf{0.78} & 21.993 & 0.630 \\
        \bottomrule
    \end{tabular}}
    \end{center}
    \vspace{-0.8cm}
\end{table}

%% file: sec/6_conclusion.tex
\section{Conclusion}
\label{sec:conclusion}

We present MapReduce LoRA and RaTE, two complementary methods to address the "alignment tax" in multi-preference post-training. By iteratively merging parallel reward experts and distilling them into composable token embeddings, our framework converts objective trade-offs into a controllable tuning strategy. Extensive evaluations across text-to-image (SD 3.5, FLUX.1), video (HunyuanVideo), and language (Llama-2) models demonstrate substantial gains in both targeted and untargeted metrics. Overall, our approach offers a simple, scalable recipe for systematically advancing the multi-preference Pareto front and enabling practical model customization.


%% file: sec/7_ack.tex
\section{Acknowledgment}

We appreciate the valuable suggestions provided by Tsung-Han Wu, Yu-Heng Hung, Fengzhe Zhou, and Jitesh Jain for this paper. This research was supported in part by the National Science Foundation under Award \#2427478 - CAREER Program, and by the National Science Foundation and the Institute of Education Sciences, U.S. Department of Education under Award \#2229873 - National AI Institute for Exceptional Education. This project was also partially supported by cyberinfrastructure resources and services provided by College of Computing at the Georgia Institute of Technology, Atlanta, Georgia, USA.

%% file: sec/X_supp.tex
\section*{\centering Supplementary Materials}
\renewcommand{\numberline}[1]{#1.\enspace\space}
\startcontents[supp]
\printcontents[supp]{l}{1}{%
}{}

\clearpage

\input{sec/supp/0_llm_task}
\input{sec/supp/1_pseudocode}

\input{sec/supp/2_implement_details}

\input{sec/supp/3_ParetoFront_FullResults}
\input{sec/supp/4_PF_comp_MORL_MapReduceLoRA}
\input{sec/supp/4_1_scalability_to_5_rewards}
\input{sec/supp/5_more_qrs}
\input{sec/supp/6_limitations}

%% file: sec/supp/0_llm_task.tex
\section{MapReduce LoRA on Language Tasks}
\label{supp:sec:language_tasks}

\begin{figure*}[h]
    \centering
    \begin{minipage}[b]{0.48\linewidth}
        \centering
        \includegraphics[width=\linewidth]{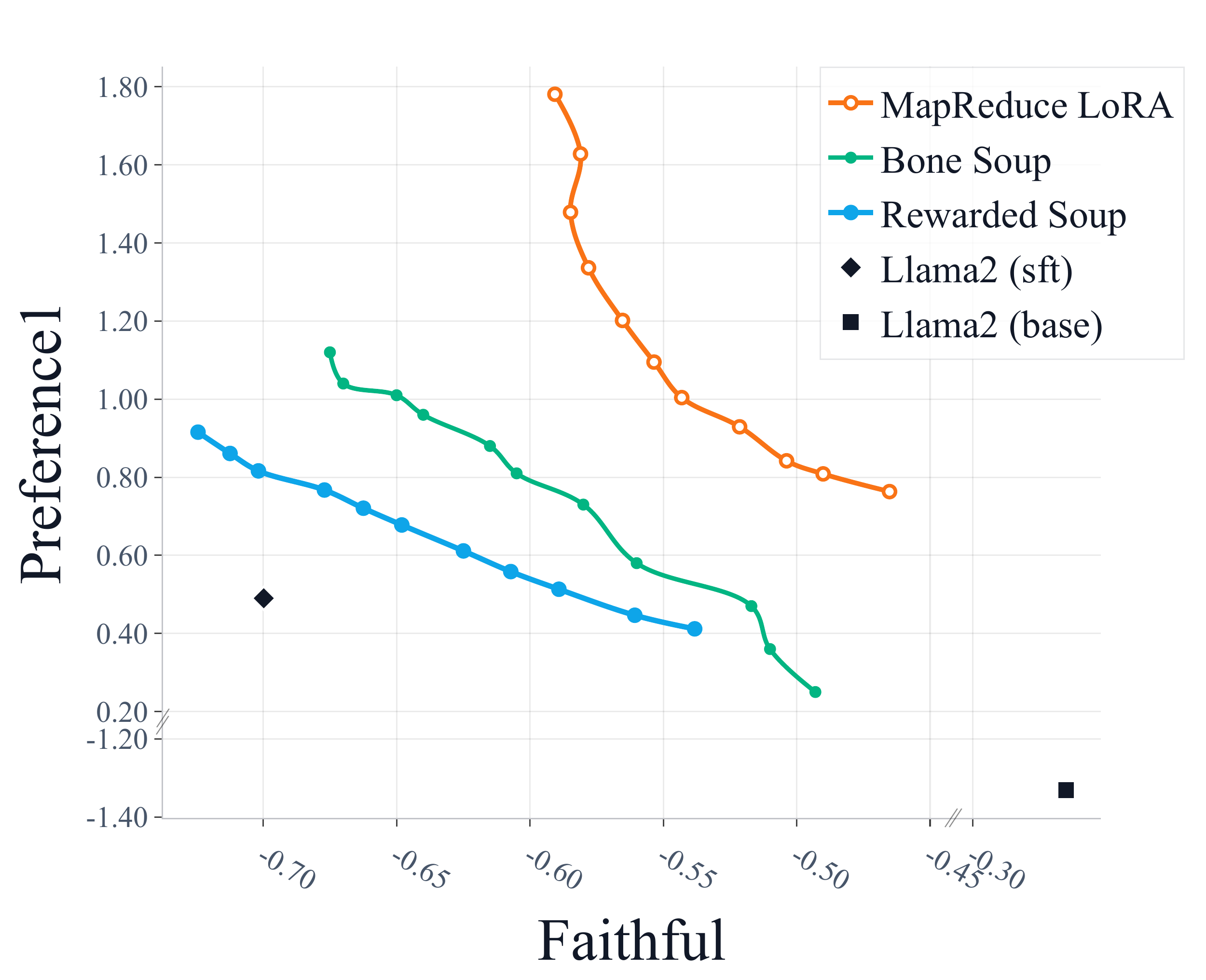}
    \end{minipage}\hfill
    \begin{minipage}[b]{0.48\linewidth}
        \centering
        \includegraphics[width=\linewidth]{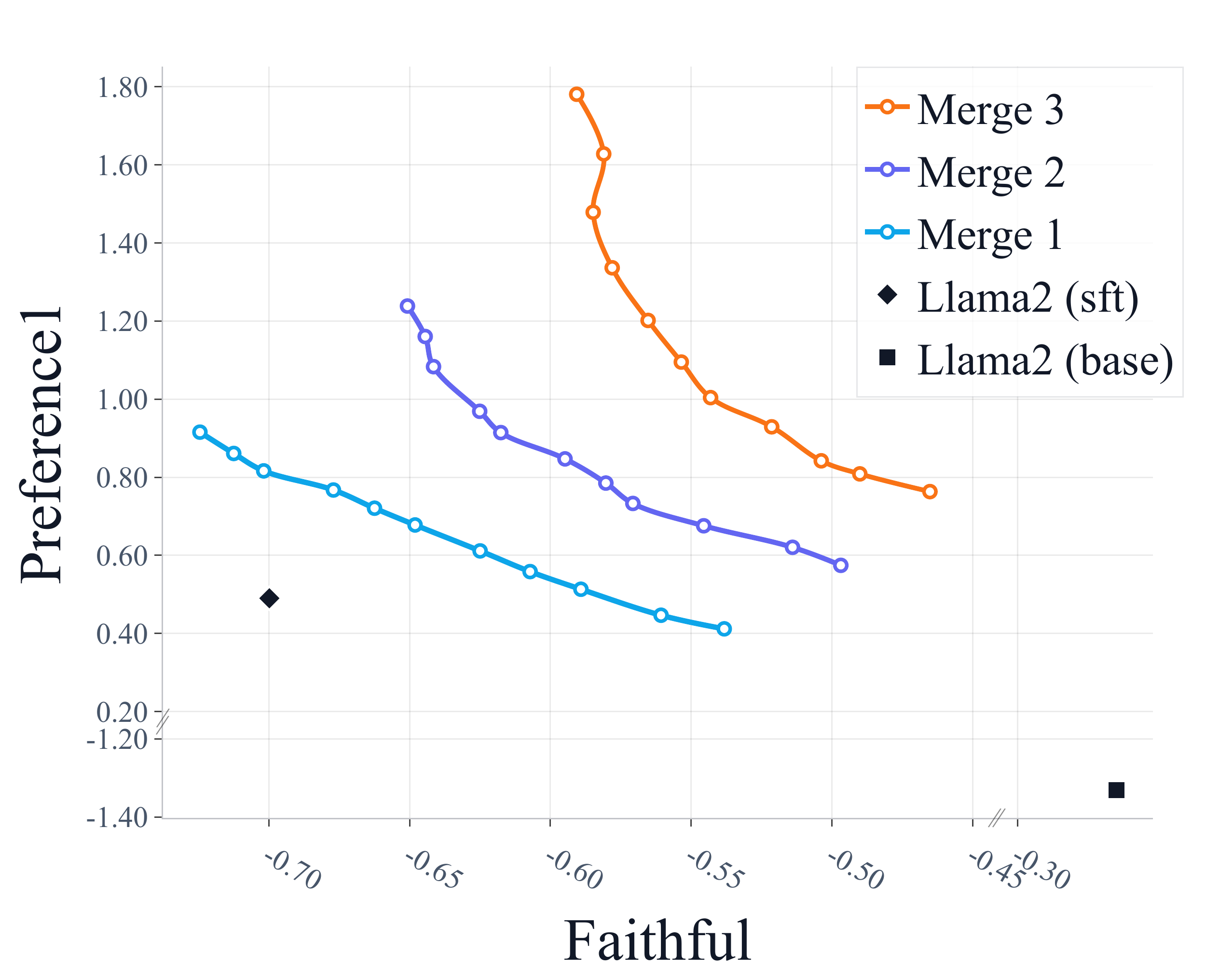}
    \end{minipage}
    \caption{\textbf{Language Task: Reddit Summary.} Left: MapReduce LoRA outperforms Rewarded Soup~\cite{reward_soup_nips'23} and Bone Soup~\cite{bonesoup_acl25} (reproduced from Fig.~5(c) in \cite{bonesoup_acl25}) on both rewards. Right: MapReduce LoRA improves progressively across merge iterations.}
     \label{supp:fig:reddit_summary_pf_llm}
\end{figure*}

\begin{figure*}[h]
    \centering
    \begin{minipage}[b]{0.48\linewidth}
        \centering
        \includegraphics[width=\linewidth]{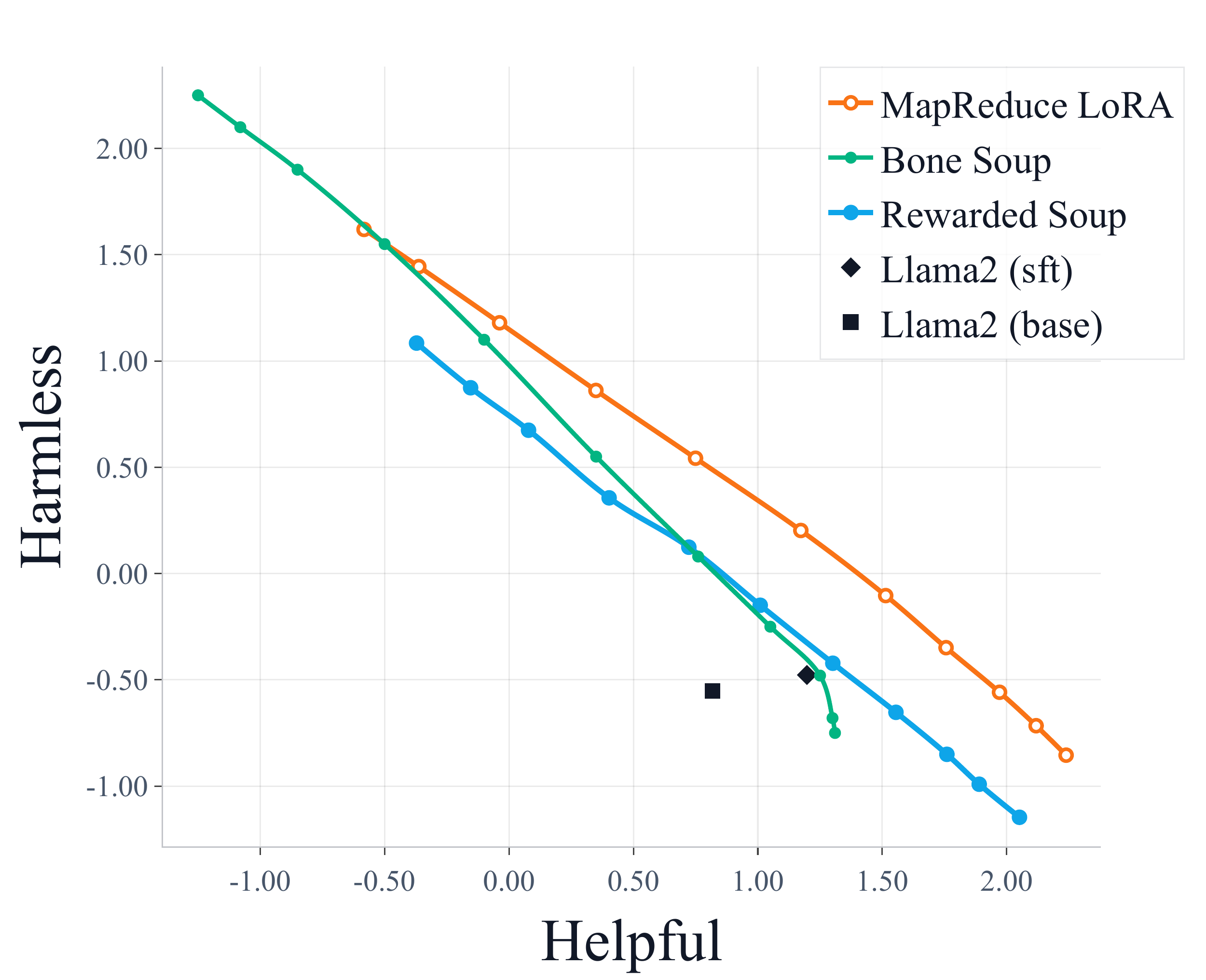}
    \end{minipage}\hfill
    \begin{minipage}[b]{0.48\linewidth}
        \centering
        \includegraphics[width=\linewidth]{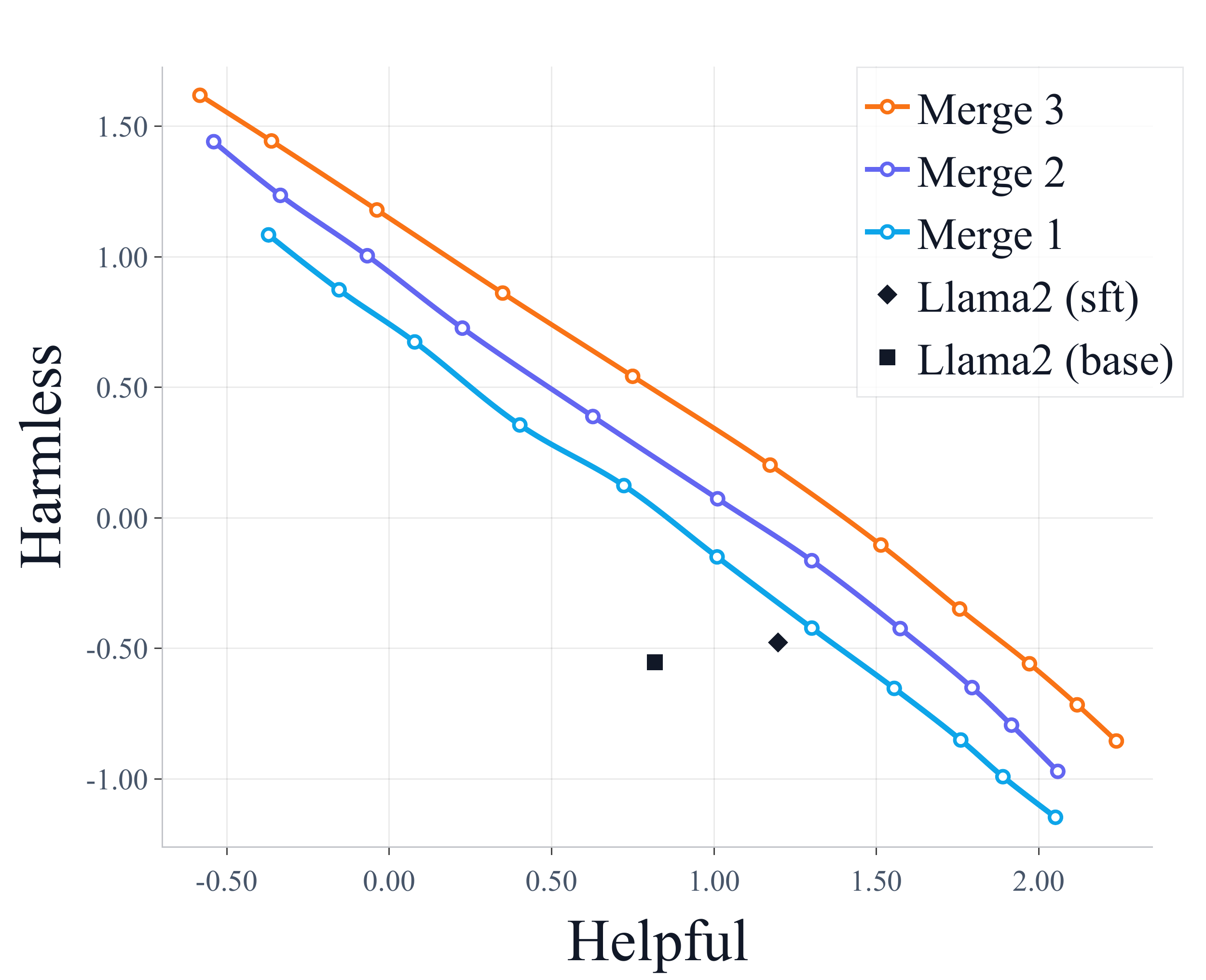}
    \end{minipage}
    \caption{\textbf{Language Task: Helful Assistant.} Left: MapReduce LoRA outperforms Rewarded Soup~\cite{reward_soup_nips'23} and Bone Soup~\cite{bonesoup_acl25} (reproduced from Fig.~5(a) in \cite{bonesoup_acl25}) on both rewards. Right: MapReduce LoRA improves progressively across merging iterations.}
     \label{supp:fig:assistant_pf_llm}
\end{figure*}

Beyond Text-to-Image and Text-to-Video, we evaluate MapReduce LoRA on language tasks to demonstrate cross-modal generality. Following the Bone Soup~\cite{bonesoup_acl25} setup, we consider two tasks, each with two rewards: (i) Reddit Summary (Faithful, Preference1) and (ii) Helpful Assistant (Helpful, Harmless). We first train the Llama2-7B~\cite{llama2_arxiv23} backbone via supervised fine-tuning (SFT) using the same datasets and hyperparameters as Bone Soup~\cite{bonesoup_acl25}, then train LoRA experts with PPO~\cite{ppo_arxiv17} under the same settings of datasets and training hyperparameters. In Figs.~\ref{supp:fig:reddit_summary_pf_llm} and \ref{supp:fig:assistant_pf_llm}, we compare MapReduce LoRA against Llama2 (base)~\cite{llama2_arxiv23}, Llama2 after SFT, Rewarded Soup~\cite{reward_soup_nips'23}, and Bone Soup~\cite{bonesoup_acl25}. MapReduce LoRA achieves state-of-the-art performance on both tasks, underscoring its cross-modal generality.





%% file: sec/supp/1_pseudocode.tex
\section{Pseudocode}
\label{supp:sec:pseudocode}

\subsection{Train: MapReduce LoRA}
\label{supp:sec:pseudocode:mapreduce-lora}
\vspace{-.4cm}
\begin{algorithm}[H]
    \caption{MapReduce LoRA: Multi-Preference Training}
    \label{alg:mapreduce-lora}
    \begin{algorithmic}[1]
    
    \Require Base model $M$ with parameters $\theta^{(0)}$; Reward models $\{R_i\}_{i=1}^n$; LoRA configuration (target layers, rank $r$, scale $\alpha$); Number of iterations $K$; GRPO steps $T_{\text{GRPO}}$; Merge weights $\{\mu_i\}_{i=1}^n$ (default $\mu_i = 1/n$, s.t. $\mu_i \ge 0$ and $\sum_{i=1}^{n} \mu_i = 1$); GRPO hyperparameters: Group size $G$, Clip $\epsilon$, KL weight $\beta$
    
    \Ensure Multi-preference aligned model $M^{(K)}$
    
    \State Initialize reference policy $\pi_{\text{ref}} \gets M(\theta^{(0)})$
    \For{$k = 0$ to $K-1$}
        \ForAll{$i \in \{1,\dots,n\}$} \Comment{Map phase: train per-reward LoRA experts in parallel}
            \State Initialize LoRA adapter $\phi_i^{(k)}$
            \State Freeze base weights $\theta^{(k)}$; only $\phi_i^{(k)}$ is trainable
            \For{$t = 1$ to $T_{\text{GRPO}}$}
                \State $\phi_i^{(k)} \gets \text{GRPO}\big(
                M, \theta^{(k)}, \phi_i^{(k)}, R_i, \pi_{\text{ref}},
                G, \epsilon, \beta \big)$
            \EndFor
        \EndFor
        \State $\bar{\phi}^{(k)} \gets \sum_{i=1}^{n} \mu_i \, \phi_i^{(k)}$ \Comment{Reduce phase: average experts}
        \State $\theta^{(k+1)} \gets \text{MergeLoRAIntoBase}\big(\theta^{(k)}, \bar{\phi}^{(k)}\big)$ \Comment{update base model}
        \State $M \gets M(\theta^{(k+1)})$
        \State Reset all adapters $\phi_i^{(k)}$ to zero
        \State Set $\pi_{\text{ref}} \gets M(\theta^{(k+1)})$
    \EndFor
    \State \Return $M^{(K)}$
    \end{algorithmic}
\end{algorithm}
\vspace{-0.25cm}

\subsection{Train: Reward-aware Token Embedding}
\label{supp:sec:pseudocode:reward-aware-token-embedding}
\vspace{-.4cm}
\begin{algorithm}[H]
    \caption{Reward-aware Token Embedding (RaTE) Training}
    \label{alg:rate}
    \begin{algorithmic}[1]
    \Require Frozen base model $M$ with parameters $\theta$;
    per-reward LoRA expert $\phi_i$ (teacher) for reward $R_i$;
    training prompts $\mathcal{D}$; special token index $RaTE_i$ for reward $i$;
    number of RaTE steps $T_{\text{RaTE}}$; Flow Matching noise schedule $\{\sigma_t\}$
    \Ensure Trained token embedding $\theta^{\text{token}}_i$ for reward $i$
    \State Attach LoRA expert $\phi_i$ to $M$ to form teacher model $M_{\text{teacher}}$
    \State Freeze all Transformer weights and all token embeddings except $\theta^{\text{token}}_i$
    \For{$t = 1$ {\bf to} $T_{\text{RaTE}}$}
        \State Sample a minibatch of prompts $\{p_b\}_{b=1}^B$ from $\mathcal{D}$
        \For{each prompt $p$ in $\{p_b\}$}
            \State // \textbf{Teacher: obtain preference-specific latent}
            \State Sample initial noise $\epsilon_0 \sim \mathcal{N}(0, I)$
            \State Generate teacher latent $z^{\text{teacher}}_{0,i} \leftarrow 
                \textsc{Sample}(M_{\text{teacher}}, p, \epsilon_0)$
            \State // \textbf{Flow Matching distillation setup}
            \State Sample $t$ via a flow-matching schedule (e.g., logit-normal over timesteps); sample $\epsilon \sim \mathcal{N}(0, I)$
            \State Compute $\sigma_{t}$ from the scheduler
            \State $z_{t} \leftarrow (1-\sigma_{t}) \, z^{\text{teacher}}_{0,i} + \sigma_{t} \, \epsilon$
            \State Target velocity: $v_{\text{target}} \leftarrow \epsilon - z^{\text{teacher}}_{0,i}$
            \State // \textbf{Student: base model + special token}
            \State Construct prompt: $p' = \text{Concat}(p, \texttt{\string<}RaTE_i\texttt{\string>})$
            \State Student velocity prediction: $v_{\text{pred}} \leftarrow M(z_{t}, t, c(p', \theta^{\text{token}}_i))$
            \State Loss: $\mathcal{L}_{\text{RaTE}} \leftarrow \big\| v_{\text{pred}} - v_{\text{target}} \big\|_2^2$
            \State Backpropagate and update only $\theta^{\text{token}}_i$: $\theta^{\text{token}}_i \leftarrow \theta^{\text{token}}_i - \eta_{\text{RaTE}} \, \nabla_{\theta^{\text{token}}_i} \mathcal{L}_{\text{RaTE}}$
        \EndFor
    \EndFor
    \State \Return $\theta^{\text{token}}_i$
    \end{algorithmic}
\end{algorithm}

\subsection{Inference with MapReduce LoRA and RaTE}
\label{supp:sec:pseudocode:inference}
\vspace{-.4cm}
\begin{algorithm}[H]
    \caption{Inference with MapReduce LoRA and RaTE}
    \label{alg:inference}
    \begin{algorithmic}[1]
    \Require Final merged model $M^{(K)}$ with parameters $\theta^{(K)}$;
    reward-aware tokens $\{\texttt{\string<}RaTE_i\texttt{\string>}\}$ and embeddings $\{\theta^{\text{token}}_i\}$;
    user prompt $p$; user-specified preference set $\mathcal{S} \subseteq \{1,\dots,n\}$;
    sampling hyperparameters (number of steps, seeds, guidance scale, etc.)
    \Ensure Generated image or video sample $x$
    \State Build preference-aware prompt:
    \State \quad $p^* \leftarrow p$
    \For{each $i \in \mathcal{S}$ in user-defined order}
        \State Append control token: $p^* \leftarrow \text{Concat}(p^*, \texttt{\string<}RaTE_i\texttt{\string>})$
    \EndFor
    \State Sample initial noise (image or video latent) $\epsilon_0 \sim \mathcal{N}(0, I)$
    \State Run the sampler with $M^{(K)}$ conditioned on $p^*$:
    \State \quad $x \leftarrow \textsc{Sample}(M^{(K)}, p^*, \epsilon_0)$
    \State \Return $x$
    \end{algorithmic}
\end{algorithm}

%% file: sec/supp/2_implement_details.tex
\section{Implementation Details}
\label{supp:sec:implementation_details}
\subsection{Training Configuration}

\paragraph{Reward-aware Token Embedding.} For each reward-aware token embedding, it contains 3 embeddings corresponding to 3 text encoders within SD 3.5 M~\cite{sd35m_hf}. The embedding is trained on the GenEval, PickScore, and OCR datasets, individually. The teacher model is trained with LoRA via GRPO~\cite{grpo_arxiv'24} with 4,200 steps on GenEval~\cite{GenEval_NeurIPS'23}, 4,500 steps on PickScore~\cite{PickScore_NeurIPS'23} and 1,600 steps on OCR~\cite{paddleocr_arxiv'25}. The training batchsize for GenEval is 512, PickScore is 768 and OCR is 768. We use the AdamW optimizer~\cite{adamw_iclr'19} with learning rate 3e-4, $\beta_1$ = 0.9, $\beta_2$ = 0.999, weight decay = 1e-4 and no warmup schedule; no KL penalty is applied.

\paragraph{MapReduce LoRA.} 
We adopt uniform averaging for all the default experiments if not specified. 

\begin{itemize}
    \item \textbf{Text-to-Image}: We train all models using AdamW optimizer~\cite{adamw_iclr'19} with learning rate 3e-4, $\beta_1$ = 0.9, $\beta_2$ = 0.999, weight decay = 1e-4, and no warmup schedule. All experiments use 32 GPUs with global batch size 576.

    \begin{itemize}[label=$\circ$]
        \item \textbf{SD 3.5 M~\cite{sd35m_hf}}: We follow the training configuration in Flow-GRPO~\cite{flow_grpo_arxiv'25}, where only the KL ratio $\beta$ is different. $\beta_{\text{GenEval}}$ = 0.04, $\beta_{\text{PickScore}}$ = 0.01, and $\beta_{\text{OCR}}$ = 0.04. The sampling timestep $T$ is 10 and the evaluation timestep $T$ = 40. The GRPO group size $G$ is 24, the resolution is 512, and the LoRA settings are $\alpha$ = 64 and $r$ = 32. LoRA is applied to all attention layers (\texttt{attn.add\_q\_proj}, \texttt{attn.add\_k\_proj}, \texttt{attn.add\_v\_proj}, \texttt{attn.to\_add\_out}, \texttt{attn.to\_q}, \texttt{attn.to\_k}, \texttt{attn.to\_v}, \texttt{attn.to\_out.0}). The per-GPU batch sizes are 6, 9, and 9 with gradient accumulation steps of 3, 2, and 2 for GenEval, PickScore, and OCR, respectively. We train GenEval, PickScore and OCR for \{4,100, 1,200, 1,000, 1,100\}, \{4,500, 1,400, 1,000, 1,000\}, and \{1,600, 1,300, 900, 1,100\} steps, respectively; the bracketed numbers denote merge iterations 0-3.
        \item \textbf{FLUX.1-dev~\cite{flux24}}: The only task-dependent change is the KL ratio $\beta$ ($\beta_{\text{GenEval}}{=}0.04$, $\beta_{\text{PickScore}}{=}0$, $\beta_{\text{OCR}}{=}0.04$). The sampling timestep $T$ is 6 and the evaluation timestep $T$ = 28. The GRPO group size $G$ is 24, the resolution is 512, and the LoRA settings are $\alpha$ = 128 and $r$ = 64. LoRA is applied to attention layers and feed-forward network layers (\texttt{attn.to\_q}, \texttt{attn.to\_k}, \texttt{attn.to\_v}, \texttt{attn.to\_out.0}, \texttt{attn.add\_q\_proj}, \texttt{attn.add\_k\_proj}, \texttt{attn.add\_v\_proj}, \texttt{attn.to\_add\_out}, \texttt{ff.net.0.proj}, \texttt{ff.net.2}, \texttt{ff\_context.net.0.proj}, \texttt{ff\_context.net.2}). The per-GPU batch size is 3 with gradient accumulation steps of 6 for all three tasks. We train GenEval, PickScore and OCR for \{2,700, 1,900, 1,800, 1,600\}, \{1,250, 600, 1,550, 2,200\}, and \{1,250, 650, 850, 1,100\} steps, respectively; the bracketed numbers denote merge iterations 0-3.
    \end{itemize}

    \item \textbf{Text-to-Video}: We train the HunyuanVideo~\cite{hunyuanvideo_arxiv'25} with the same configuration as the DanceGRPO~\cite{dancegrpo_arxiv25} but apply LoRA instead of full finetuning. The LoRA settings are $\alpha$ = 64 and $r$ = 32. LoRA is applied to attention layers ``\texttt{attn.to\_q}, \texttt{attn.to\_k}, \texttt{attn.to\_v}'' in single\_transformer\_blocks (40 blocks), ``\texttt{attn.to\_q}, \texttt{attn.to\_k}, \texttt{attn.to\_v}, \texttt{attn.\allowbreak to\_out.0}, \texttt{attn.to\_add\_out}, \texttt{attn.add\_q\_proj},  \texttt{attn.add\_k\_proj}, \texttt{attn.add\_v\_proj}'' in transformer\_blocks (20 blocks). We train VQ and MQ for \{200, 140, 100\}, \{195, 145, 115\} steps, respectively; the bracketed numbers denote merge iterations 0-2.

\end{itemize}

\paragraph{Multi-objective Reinforcement Learning (MORL).} 
The hyperparameters for MORL are identical to the MapReduce LoRA mentioned above. During each epoch, batches are sampled from GenEval~\cite{GenEval_NeurIPS'23}, PickScore~\cite{PickScore_NeurIPS'23}, and OCR~\cite{paddleocr_arxiv'25} datasets according to specified ratios (default 1:1:1), where each individual batch contains samples from a single source. During evaluation, each data source is independently evaluated using its corresponding reward model to provide per-task performance metrics. The training steps are comparable to the overall training steps of MapReduce LoRA for fairness. There are two variations of MORL in our experiments, and the key difference lies in the data mixture and reward mixture strategy: 
\begin{itemize}
    \item \textbf{Data mixture (MORL-D)}: Each sample is only evaluated by its corresponding reward model based on its source dataset.
    \item \textbf{Data mixture and reward mixture (MORL-DR)}: Samples from GenEval~\cite{GenEval_NeurIPS'23} and OCR~\cite{paddleocr_arxiv'25} are scored by both their task-specific reward and PickScore~\cite{PickScore_NeurIPS'23}; we take the average as the final reward (e.g., for GenEval, $(r_{\text{geneval}} + r_{\text{pickscore}})/2$). In contrast, samples from the PickScore dataset are evaluated only with PickScore. This choice reflects dataset constraints: GenEval requires a structured prompt format, whereas OCR requires prompts that contain text for a valid evaluation. PickScore serves as a shared aesthetic quality anchor across tasks. Overall, this protocol better aligns with the multi-objective setting by applying multiple rewards in a dataset-agnostic manner (where applicable), rather than restricting evaluation to dataset-specific rewards.
\end{itemize}

\paragraph{Compute Resources.} All the experiments are conducted on 32 NVIDIA A100 GPUs (80GB).

\subsection{Base Models and Reward Models}

The following table presents the base model and reward models along with their corresponding links.

\begin{table}[h]
    \centering
    \resizebox{\linewidth}{!}{
    \begin{tabular}{ll}
    \toprule
    \textbf{Models} & \textbf{Links} \\
    \midrule
    \texttt{SD 3.5 M}~\citep{sd35m_hf} & \url{https://huggingface.co/stabilityai/stable-diffusion-3.5-medium} \\
    \texttt{FLUX.1-dev}~\citep{flux24} & \url{https://huggingface.co/black-forest-labs/FLUX.1-dev} \\
    \texttt{HunyuanVideo}~\citep{hunyuanvideo_arxiv'25} & \url{https://huggingface.co/hunyuanvideo-community/HunyuanVideo} \\
    \bottomrule
    \end{tabular}
    }
\end{table}

\begin{table}[h]
    \centering
    \resizebox{\linewidth}{!}{
    \begin{tabular}{ll}
    \toprule
    \textbf{Reward Models} & \textbf{Links} \\
    \midrule
    \texttt{GenEval}~\citep{GenEval_NeurIPS'23} & \url{https://github.com/djghosh13/geneval} \\
    \texttt{PickScore}~\citep{PickScore_NeurIPS'23} & \url{https://huggingface.co/yuvalkirstain/PickScore_v1} \\
    \texttt{OCR}~\citep{paddleocr_arxiv'25} & \url{https://github.com/PaddlePaddle/PaddleOCR} \\
    \texttt{VQAScore}~\citep{VQAScore_ECCV24} & \url{https://github.com/linzhiqiu/t2v_metrics} \\
    \texttt{MPS}~\citep{MPS_CVPR'24} & \url{https://github.com/Kwai-Kolors/MPS} \\
    \texttt{VILA}~\citep{VILA_CVPR'23} & \url{https://github.com/google-research/google-research/tree/master/vila} \\
    \texttt{VQ} and \texttt{MQ}~\cite{videoalign_arxiv25} & \url{https://github.com/KlingTeam/VideoAlign/tree/main} \\
    \bottomrule
    \end{tabular}
    }
\end{table}

%% file: sec/supp/3_ParetoFront_FullResults.tex
\section{More Quantitative Results}
\subsection{Full Results of Different Merging Ratios (Corresponding to Fig.~\ref{fig:pareto_front_teaser})}
\label{supp:sec:full_results_fig1}

\begin{figure*}[h]
    \centering
    \includegraphics[width=1.0\linewidth]{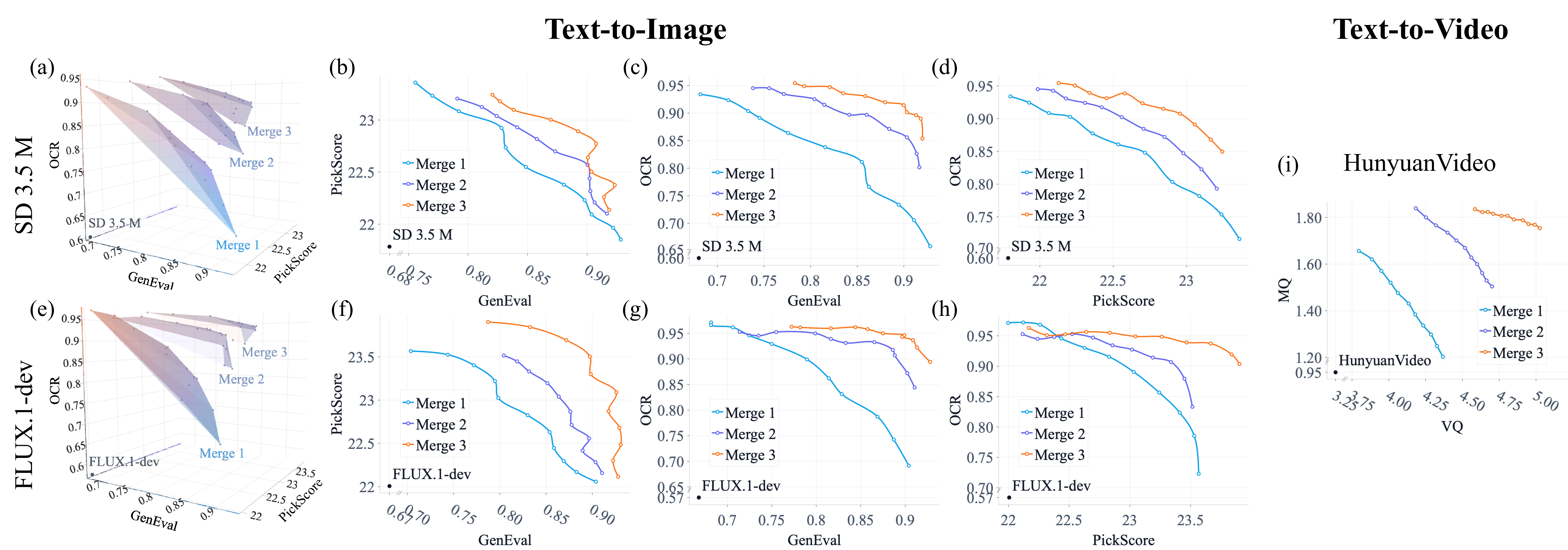}
    \vspace{-.7cm}
     \caption{\textbf{MapReduce LoRA advances the Pareto fronts on Text-to-Image and Text-to-Video tasks.} We include the non-Pareto sets in this figure for sharing the full results.}  
     \label{fig:pf_full_results}
     \vspace{-.3cm}
\end{figure*}

\textbf{Pareto fronts} denote the set of non-dominated solutions (parameter settings for which no objective can be improved without degrading another). In Fig.~\ref{fig:pf_full_results}, we include dominated (non-Pareto-optimal) points to present the full result set, together with the scores reported in Tables~\ref{tab:3D_t2i_merge_results} and \ref{tab:2D_t2i_t2v_merge_results}. Dominated points can arise for four reasons:

\begin{itemize}
    \item \textbf{Estimation noise and stochasticity}: Each reward is estimated from finite samples and noisy reward evaluators. For GenEval, PickScore and OCR, we evaluate 553, 2,048 and 1,018 cases, respectively.
    \item \textbf{Non-convex objective landscapes}: The optimization surfaces can be highly non-convex. The merged model checkpoints may lie in different basins or apexes for different reward functions.
    \item \textbf{Partial or misaligned objectives}: When the two rewards are not orthogonal (\ie, they share underlying structure), optimizing one can initially improve the other; once the shared structure is exhausted, trade-offs dominate again, yielding dominated points.
    \item \textbf{Training unsaturation}: In Fig.~\ref{fig:pf_full_results}~(g) and (h), the edge points are mixed together on the top-left corner because iteration 1 for OCR uses much fewer training steps (650) than iteration 0 (1,250); additional training may further improve performance.
\end{itemize}

\begin{wrapfigure}[35]{r}{0.65\textwidth}
    \vspace{-0.4cm}
    \begin{center}
    \small
    \makeatletter\def\@captype{table}\makeatother
    \caption{3D merging results on Text-to-Image tasks: SD 3.5 M~\cite{sd35m_hf} and FLUX.1-dev~\cite{flux24}.}
    \label{tab:3D_t2i_merge_results}
    \vspace{-0.3cm}
    \scalebox{0.75}{%
    \setlength{\tabcolsep}{0.3em}
    \begin{tabular}{lccc|ccc|ccc}
        \toprule
        \multirow{2}{*}{Group} & \multicolumn{3}{c}{Merging Ratios} & \multicolumn{3}{c}{SD 3.5 M} & \multicolumn{3}{c}{FLUX.1-dev} \\
        \cmidrule(lr){2-4}\cmidrule(lr){5-7}\cmidrule(lr){8-10}
        & GenEval & PickScore & OCR  & GenEval & PickScore & OCR & GenEval & PickScore & OCR \\
        \midrule
        \rowcolor[HTML]{EFEFEF} 
        base model & 0 & 0 & 0 & 0.68 &	21.784 &	0.601  & 0.67 &	22.006 &	0.573\\
        \midrule
        \multirow{13}{*}{Merge 1} & 1 & 0 & 0 & 0.93 &	21.852 &	0.658 & 0.90 &	22.057 &	0.692\\
        & 0 & 1 & 0 & 0.76 &	23.359 &	0.716 & 0.70 &	23.566 &	0.723 \\
        & 0 & 0 & 1 & 0.68 &	21.800 &	0.934 & 0.68 &	21.997 &	0.971 \\
        & 0.8 & 0.1 & 0.1 & 0.90 &	21.993 &	0.704 & 0.88 &	22.175 &	0.756\\
        & 0.1 & 0.8 & 0.1 & 0.77 &	23.090 &	0.746 & 0.75 &	23.408 &	0.794\\
        & 0.1 & 0.1 & 0.8 & 0.70 &	21.948 &	0.911 & 0.71 &	22.136 &	0.959\\
        & 0.6 & 0.2 & 0.2 & 0.86 &	22.106 &	0.743 & 0.83 &	22.299 &	0.802 \\
        & 0.2 & 0.6 & 0.2 & 0.78 &	22.731 &	0.781 & 0.76 &	23.046 &	0.827\\
        & 0.2 & 0.2 & 0.6 & 0.76 &	22.087 &	0.882 & 0.73 &	22.288 &	0.928\\
        & 0.4 & 0.4 & 0.2 & 0.81 &	22.382 &	0.758 & 0.81 &	22.633 &	0.813\\
        & 0.4 & 0.2 & 0.4 & 0.80 &	22.098 &	0.826 & 0.79 &	22.294 &	0.883\\
        & 0.2 & 0.4 & 0.4 & 0.77 &	22.374 &	0.836 & 0.76 &	22.636 &	0.886\\
        \rowcolor[HTML]{D6FFCF} 
        \cellcolor{white} & $0.\overline{3}$ & $0.\overline{3}$ & $0.\overline{3}$ & 0.79 &	22.276 &	0.805 & 0.77 &	22.503 &	0.868\\
        \midrule
        \multirow{13}{*}{Merge 2} & 1 & 0 & 0 & 0.92 &	22.101 &	0.802 & 0.91 &	22.156 &	0.844\\
        & 0 & 1 & 0 & 0.79 &	23.205 &	0.793 & 0.80 &	23.515 &	0.833 \\
        & 0 & 0 & 1 & 0.74 &	21.986 &	0.945 & 0.71 &	22.116 &	0.953 \\
        & 0.8 & 0.1 & 0.1 & 0.90 &	22.208 &	0.831 & 0.89 &	22.303 &	0.883 \\
        & 0.1 & 0.8 & 0.1 & 0.82 &	23.004 &	0.820 & 0.81 &	23.337 &	0.888 \\
        & 0.1 & 0.1 & 0.8 & 0.76 &	22.109 &	0.933 & 0.75 &	22.249 &	0.956 \\
        & 0.6 & 0.2 & 0.2 & 0.89 &	22.316 &	0.839 & 0.88 &	22.437 &	0.913 \\
        & 0.2 & 0.6 & 0.2 & 0.84 &	22.778 &	0.843 & 0.83 &	23.037 &	0.916 \\
        & 0.2 & 0.2 & 0.6 & 0.81 &	22.232 &	0.917 & 0.80 &	22.405 &	0.941 \\
        & 0.4 & 0.4 & 0.2 & 0.85 &	22.532 &	0.850 & 0.85 &	22.718 &	0.918 \\
        & 0.4 & 0.2 & 0.4 & 0.84 &	22.290 &	0.895 & 0.85 &	22.429 &	0.927\\
        & 0.2 & 0.4 & 0.4 & 0.83 &	22.486 &	0.899 & 0.81 &	22.709 &	0.929\\
        \rowcolor[HTML]{D6FFCF} 
        \cellcolor{white} & $0.\overline{3}$ & $0.\overline{3}$ & $0.\overline{3}$ & 0.84 &	22.436 &	0.871 & 0.84 &	22.619 &	0.926\\
        \midrule
        \multirow{13}{*}{Merge 3} & 1 & 0 & 0 & 0.92 &	22.138 &	0.854 & 0.93 &	22.115 &	0.894\\
        & 0 & 1 & 0 & 0.82 &	23.243 &	0.850 & 0.79 &	23.902 &	0.904 \\
        & 0 & 0 & 1 & 0.78 &	22.128 &	0.954 & 0.77 &	22.167 &	0.963 \\
        & 0.8 & 0.1 & 0.1 & 0.92 &	22.281 &	0.890 & 0.91 &	22.324 &	0.928 \\
        & 0.1 & 0.8 & 0.1 & 0.83 &	23.080 &	0.875 & 0.85 &	23.683 &	0.930 \\
        & 0.1 & 0.1 & 0.8 & 0.81 &	22.248 &	0.944 & 0.80 &	22.331 &	0.960 \\
        & 0.6 & 0.2 & 0.2 & 0.91 &	22.408 &	0.900 & 0.90 &	22.508 &	0.936 \\
        & 0.2 & 0.6 & 0.2 & 0.85 &	22.869 &	0.887 & 0.87 &	23.285 &	0.938 \\
        & 0.2 & 0.2 & 0.6 & 0.84 &	22.369 &	0.929 & 0.85 &	22.496 &	0.955 \\
        & 0.4 & 0.4 & 0.2 & 0.88 &	22.637 &	0.904 & 0.90 &	22.878 &	0.934 \\
        & 0.4 & 0.2 & 0.4 & 0.87 &	22.393 &	0.920 & 0.88 &	22.515 &	0.943 \\
        & 0.2 & 0.4 & 0.4 & 0.85 &	22.615 &	0.913 & 0.86 &	22.864 &	0.946 \\
        \rowcolor[HTML]{D6FFCF} 
        \cellcolor{white} & $0.\overline{3}$ & $0.\overline{3}$ & $0.\overline{3}$ & 0.88 &	22.554 &	0.908 & 0.88 &	22.740 &	0.941 \\
        \bottomrule
    \end{tabular}}
    \end{center}
\end{wrapfigure}

\noindent\textbf{Evaluation details:}

\begin{itemize}
    \item \textbf{3D Pareto fronts}: For each merge, we evaluate thirteen model weight configurations with different coefficient ratios: three single-reward configurations (the other two coefficients set to zero) and ten mixed configurations. The specific \{GenEval: PickScore: OCR\} ratios are $\{1:0:0\}$, $\{0:1:0\}$, $\{0:0:1\}$, $\{0.8:0.1:0.1\}$, $\{0.1:0.8:0.1\}$, $\{0.1:0.1:0.8\}$, $\{0.6:0.2:0.2\}$, $\{0.2:0.6:0.2\}$, $\{0.2:0.2:0.6\}$, $\{0.4:0.4:0.2\}$, $\{0.4:0.2:0.4\}$, $\{0.2:0.4:0.4\}$, and $\{ 0.\overline{3}:0.\overline{3}:0.\overline{3} \}$. Each plotted point is obtained by evaluating one merged model on all three metrics. 
    \item \textbf{2D Pareto fronts}: For each merge, we evaluate eleven model weight configurations with different ratios: two single-reward configurations (the other coefficient set to zero) and nine mixed configurations. The specific ratios are $\{1:0\}$, $\{0:1\}$, $\{0.1:0.9\}$, $\{0.2:0.8\}$, $\{0.3:0.7\}$, $\{0.4:0.6\}$, $\{0.5:0.5\}$, $\{0.6:0.4\}$, $\{0.7:0.3\}$, $\{0.8:0.2\}$, and $\{0.9:0.1\}$. Each point is obtained by evaluating one merged model on the two metrics.
\end{itemize}

\begin{table*}[h]
    \begin{center}
    \small
    \caption{2D merging results on Text-to-Image tasks, SD 3.5 M~\cite{sd35m_hf} and FLUX.1-dev~\cite{flux24}, and Text-to-Video task, HunyuanVideo~\cite{hunyuanvideo_arxiv'25}.}
    \label{tab:2D_t2i_t2v_merge_results}
    \vspace{-0.3cm}
    \scalebox{0.73}{%
    \setlength{\tabcolsep}{0.3em}
    \begin{tabular}{lcc|cc|cc|cc|cc|cc|cc|cc}
        \toprule
        \multirow{2}{*}{Group} & \multicolumn{2}{c}{Merging Ratios} & \multicolumn{6}{c}{SD 3.5 M} & \multicolumn{6}{c}{FLUX.1-dev} & \multicolumn{2}{c}{HunyuanVideo} \\
        \cmidrule(lr){2-3}\cmidrule(lr){4-9}\cmidrule(lr){10-15}\cmidrule(lr){16-17}
        & reward A & reward B & $^A$GenEval & $^B$PickScore & $^A$GenEval & $^B$OCR & $^A$PickScore & $^B$OCR & $^A$GenEval & $^B$PickScore & $^A$GenEval & $^B$OCR & $^A$PickScore & $^B$OCR & $^A$VQ & $^B$MQ \\
        \midrule
        \rowcolor[HTML]{EFEFEF} 
        base model & 0 & 0 & 0.68 &	21.784 &	0.68 &	0.601 &	21.784 &	0.601 & 0.67 &	22.006 & 0.67 &	0.573 & 22.006 &	0.573 & 3.25 &	0.95\\
        \midrule
        & 0 & 1 & 0.76 &	23.359 & 0.68 &	0.934 & 21.800 &	0.934 & 0.70 &	23.566 & 0.68 &	0.971 & 21.997 &	0.971 & 3.80 &	1.66 \\
        & 0.1 & 0.9 & 0.77 &	23.233 & 0.71 &	0.923 & 21.926 &	0.925 & 0.74 &	23.525 & 0.68 &	0.966 & 22.122 &	0.972 & 3.89 &	1.62 \\
        & 0.2 & 0.8 & 0.79 &	23.084 & 0.73 &	0.904 & 22.062 &	0.909 & 0.77 &	23.403 & 0.71 &	0.962 & 22.260 &	0.968 & 3.95 &	1.57 \\
        & 0.3 & 0.7 & 0.83 &	22.923 & 0.75 &	0.891 & 22.207 &	0.903 & 0.79 &	23.220 & 0.72 &	0.946 & 22.436 &	0.946 & 4.01 &	1.52\\
        & 0.4 & 0.6 & 0.83 &	22.736 & 0.78 &	0.864 & 22.358 &	0.878 & 0.80 &	23.024 & 0.75 &	0.929 & 22.627 &	0.930 & 4.06 &	1.48 \\
        \rowcolor[HTML]{D6FFCF} 
        \cellcolor{white}{Merge 1} & 0.5 & 0.5 & 0.85 &	22.549 & 0.82 &	0.838 & 22.534 &	0.861 & 0.83 &	22.825 & 0.79 &	0.899 & 22.828 &	0.916 & 4.13 &	1.43\\
        & 0.6 & 0.4 & 0.88 &	22.379 & 0.86 &	0.811 & 22.715 &	0.848 & 0.85 &	22.629 & 0.81 &	0.862 & 23.032 &	0.891 & 4.18 &	1.38\\
        & 0.7 & 0.3 & 0.90 &	22.232 & 0.86 &	0.766 & 22.902 &	0.803 & 0.86 &	22.446 & 0.83 &	0.831 & 23.242 &	0.857 & 4.23 &	1.34\\
        & 0.8 & 0.2 & 0.90 &	22.093 & 0.89 &	0.734 & 23.085 &	0.782 & 0.87 &	22.297 & 0.87 &	0.787 & 23.410 &	0.823 & 4.29 &	1.30\\
        & 0.9 & 0.1 & 0.92 &	21.966 & 0.91 &	0.706 & 23.237 &	0.754 & 0.88 &	22.169 & 0.89 &	0.742 & 23.530 &	0.785 & 4.33 &	1.25\\
        & 1 & 0 & 0.93 &	21.852 & 0.93 &	0.658 & 23.359 &	0.716 & 0.90 &	22.057 & 0.90 &	0.692 & 23.566 &	0.723 & 4.37 &	1.20\\
        \midrule
        & 0 & 1 & 0.79 &	23.205 & 0.74 &	0.945 & 21.986 &	0.945 & 0.80 &	23.515 & 0.71 &	0.953 & 22.116 &	0.953 & 4.18 &	1.84 \\
        & 0.1 & 0.9 & 0.81 &	23.125 & 0.76 &	0.945 & 22.090 &	0.943 & 0.82 &	23.447 & 0.74 &	0.945 & 22.241 &	0.945 & 4.25 &	1.80 \\
        & 0.2 & 0.8 & 0.82 &	23.039 & 0.77 &	0.935 & 22.178 &	0.931 & 0.83 &	23.332 & 0.75 &	0.952 & 22.377 &	0.948 & 4.32 &	1.77 \\
        & 0.3 & 0.7 & 0.84 &	22.930 & 0.80 &	0.925 & 22.307 &	0.924 & 0.85 &	23.195 & 0.80 &	0.950 & 22.524 &	0.953 & 4.40 &	1.73 \\
        & 0.4 & 0.6 & 0.86 &	22.817 & 0.81 &	0.915 & 22.421 &	0.917 & 0.86 &	23.040 & 0.82 &	0.939 & 22.696 &	0.947 & 4.46 &	1.70 \\
        \rowcolor[HTML]{D6FFCF} 
        \cellcolor{white}{Merge 2} & 0.5 & 0.5 & 0.87 &	22.699 & 0.84 &	0.897 & 22.557 &	0.902 & 0.88 &	22.867 & 0.83 &	0.931 & 22.856 &	0.934 & 4.52 &	1.67\\
        & 0.6 & 0.4 & 0.90 &	22.569 & 0.86 &	0.897 & 22.706 &	0.884 & 0.88 &	22.713 & 0.87 &	0.933 & 23.017 &	0.927 & 4.56 &	1.63 \\
        & 0.7 & 0.3 & 0.90 &	22.438 & 0.88 &	0.871 & 22.849 &	0.872 & 0.90 &	22.557 & 0.89 &	0.918 & 23.189 &	0.914 & 4.60 &	1.60 \\
        & 0.8 & 0.2 & 0.90 &	22.318 & 0.90 &	0.856 & 22.977 &	0.847 & 0.89 &	22.418 & 0.89 &	0.907 & 23.339 &	0.907 & 4.63 &	1.56 \\
        & 0.9 & 0.1 & 0.91 &	22.208 & 0.91 &	0.826 & 23.100 &	0.823 & 0.90 &	22.282 & 0.90 &	0.872 & 23.450 &	0.879 & 4.66 &	1.53 \\
        & 1 & 0 & 0.92 &	22.101 & 0.92 &	0.802 & 23.205 &	0.793 & 0.91 &	22.156 & 0.91 &	0.844 & 23.515 &	0.833 & 4.70 &	1.50\\
        \midrule
        & 0 & 1 & 0.82 &	23.243 &  0.78 &	0.954 & 22.128	&	0.954 & 0.79 &	23.902 & 0.77 &	0.963 & 22.167 &	0.963 & 4.59 &	1.84 \\
        & 0.1 & 0.9 & 0.83	&	23.179 & 0.79	&	0.949 & 22.238 &	0.951 & 0.83 &	23.844 & 0.78 &	0.962 & 22.316 &	0.951 & 4.64 &	1.82\\
        & 0.2 & 0.8 & 0.84	&	23.097 & 0.82	&	0.947 & 22.338	&	0.940 & 0.87 &	23.693 & 0.82 &	0.960 & 22.467 &	0.952 & 4.67 &	1.82\\
        & 0.3 & 0.7 & 0.87	&	23.003 & 0.83	&	0.936 & 22.454	&	0.931 & 0.90 &	23.504 & 0.84 &	0.962 & 22.642 &	0.956 & 4.71 &	1.82\\
        & 0.4 & 0.6 & 0.89	&	22.892 & 0.86	&	0.931 & 22.578	&	0.939 & 0.90 &	23.299 & 0.86 &	0.958 & 22.834 &	0.955 & 4.77 &	1.81\\
        \rowcolor[HTML]{D6FFCF} 
        \cellcolor{white}{Merge 3} & 0.5 & 0.5 & 0.91 &	22.772 & 0.88 &	0.920 & 22.699 &	0.924 & 0.93 &	23.091 & 0.87 &	0.950 & 23.046 &	0.949 & 4.81 &	1.81 \\
        & 0.6 & 0.4 & 0.90 &	22.640 & 0.90 &	0.914 & 22.842 &	0.915 & 0.92 &	22.872 & 0.90 &	0.946 & 23.263 &	0.948 & 4.85 &	1.79\\
        & 0.7 & 0.3 & 0.90 &	22.504 & 0.90 &	0.902 & 22.955 &	0.908 & 0.93 &	22.680 & 0.90 &	0.943 & 23.467 &	0.939 & 4.91 &	1.79\\
        & 0.8 & 0.2 & 0.92 &	22.375 & 0.91 &	0.896 & 23.055 &	0.891 & 0.93 &	22.485 & 0.91 &	0.936 & 23.666 &	0.937 & 4.95 &	1.77\\
        & 0.9 & 0.1 & 0.91 &	22.261 & 0.92 &	0.890 & 23.164 &	0.868 & 0.92 &	22.301 & 0.91 &	0.922 & 23.838 &	0.919 & 4.99 &	1.77\\
        & 1 & 0 & 0.92	&	22.138 & 0.92 &	0.854 & 23.243	&	0.850 & 0.93 &	22.115 & 0.93 &	0.894 & 23.902	&	0.904 & 5.03 &	1.75\\
        \bottomrule

    \end{tabular}}
    \end{center}
\end{table*}

%% file: sec/supp/4_PF_comp_MORL_MapReduceLoRA.tex
\vspace{+1cm}

\subsection{Pareto Front Comparison of MORL vs. MapReduce LoRA}
\label{supp:sec:pf_comp_morlD}

\begin{figure*}[h]
    \centering
    \includegraphics[width=\linewidth]{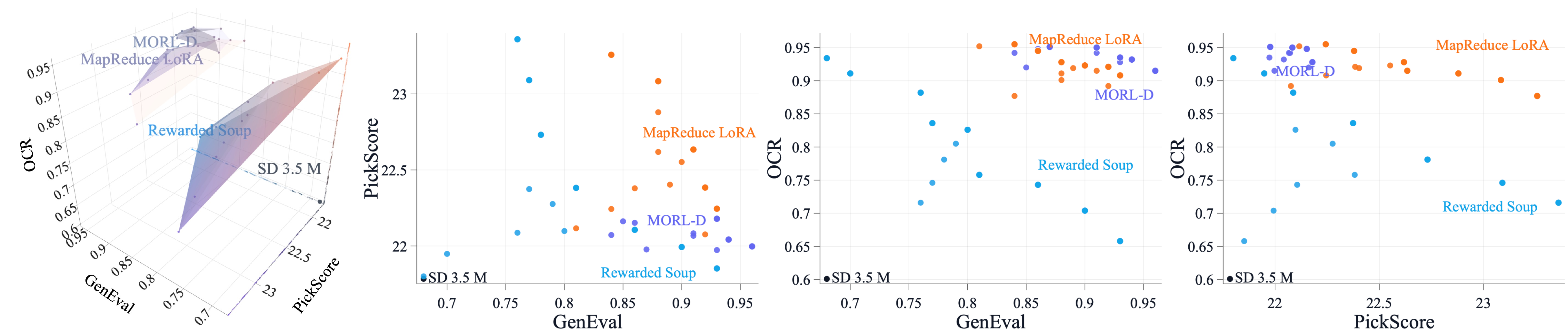}
    \caption{\textbf{Merging performance comparison across Rewarded Soup~\cite{reward_soup_nips'23}, MORL-D, and MapReduce LoRA.} Left: 3D Pareto-front comparison. Right: 2D projections of the 3D Pareto front (for readability), which are not 2D Pareto fronts. Unlike Fig.~\ref{fig:pf_full_results}, these are projections rather than a 2D merge with the third reward fixed to 0. MORL-D performance is confined to a small region across the three rewards and yields only limited improvement on PickScore.}
    \label{supp:fig:pf_comp_morlD}
\end{figure*}

To compare naive multi-objective RL with data mixing (MORL-D) against MapReduce LoRA, we train MORL-D under ten \{GenEval:\allowbreak PickScore:OCR\} sampling ratios—\{1:1:1\}, \{1:2:3\}, \{1:3:2\}, \{2:1:3\}, \{2:3:1\}, \{3:1:2\}, \{3:2:1\}, \{1:1:4\}, \{1:4:1\}, \{4:1:1\}—using the same number of training steps as MapReduce LoRA on SD 3.5 M~\cite{sd35m_hf}. Fig.~\ref{supp:fig:pf_comp_morlD} compares Pareto fronts for Rewarded Soup~\cite{reward_soup_nips'23}, MORL-D, and MapReduce LoRA (left: 3D; right: 2D projections). In 3D, MORL-D is confined to a small region across the three rewards. For readability, we also show 2D projections of the 3D front; because MORL-D is an a priori method, we cannot fix one reward to 0 as in Fig.~\ref{fig:pf_full_results}, so the right panels are projections rather than true 2D Pareto fronts. Although MORL-D is competitive on GenEval and OCR, conflicting objectives reduce visual quality (PickScore; see Fig.~\ref{fig:sd35m-comp-all-ocr}), yielding only limited PickScore gains.

%% file: sec/supp/4_1_scalability_to_5_rewards.tex
\newpage
\subsection{Scalability of Reward Numbers}
\label{supp:sec:scalability_rewards}

\begin{figure*}[h]
    \centering
    \includegraphics[width=1.0\linewidth]{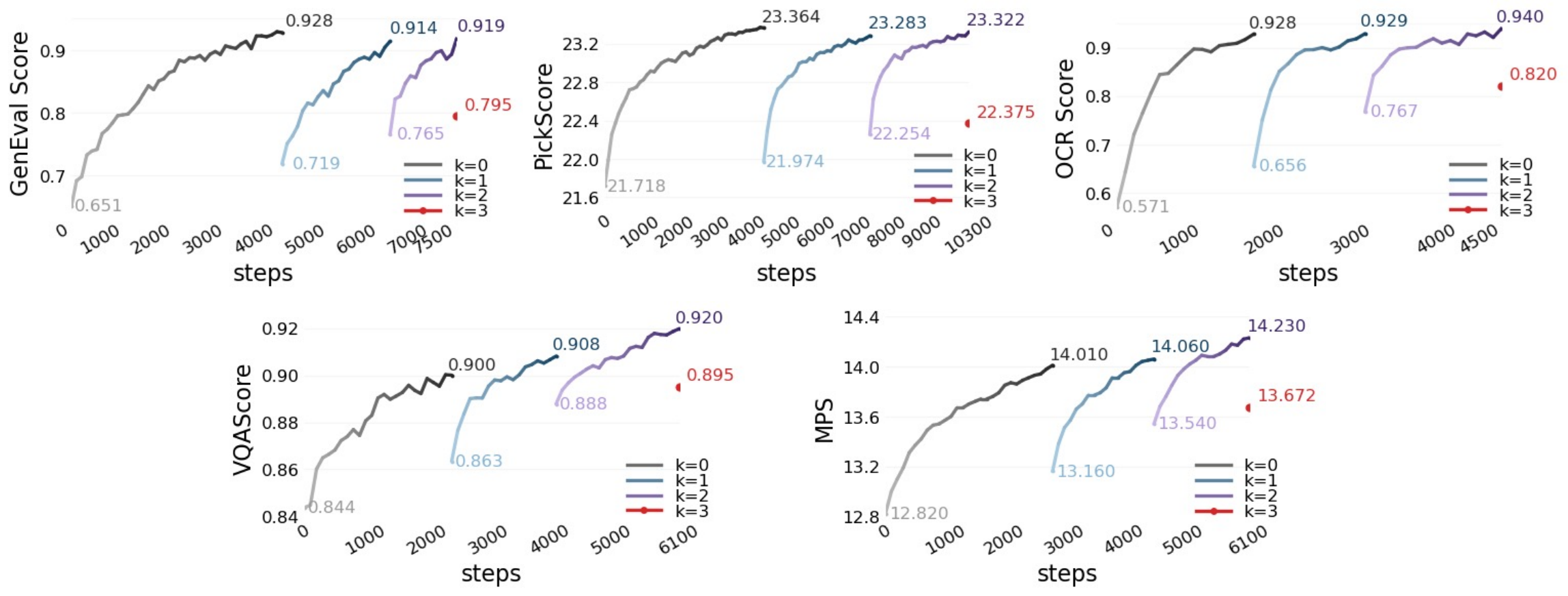}
    \vspace{-.7cm}
     \caption{MapReduce LoRA across 5 rewards with 3 merges.}  
     \label{fig:scalability_5_rewards}
     \vspace{-.3cm}
\end{figure*}

We extend MapReduce LoRA to 5 rewards by adding VQAScore~\cite{VQAScore_ECCV24} and MPS~\cite{MPS_CVPR'24} on SD 3.5 M~\cite{sd35m_hf}. Fig.~\ref{fig:scalability_5_rewards} shows that MapReduce LoRA improves all 5 rewards (+22\% GenEval, +3\% PickScore, +44\% OCR, +6\% VQAScore, +7\% MPS), consistently outperforming one-shot Rewarded Soup (the starting point of k = 1). Scaling to more rewards does not introduce diminishing returns: performance on overlapping rewards (GenEval, PickScore, OCR) remains improved compared to 3-reward training (Fig. 6, k=4). However, achieving similar levels requires ~1.66× more steps, reflecting increased optimization complexity. 
Beyond the increased optimization complexity for existing rewards, adding rewards incurs a linear per-reward training cost; however, the parallel Map phase helps mitigate the resulting wall-clock growth given sufficient GPUs.

%% file: sec/supp/5_more_qrs.tex
\newpage
\section{More Qualitative Results}
\label{supp:sec:more_qr}

\subsection{Text-to-Image qualitative results}
\label{supp:t2i_more_qual}

\begin{figure*}[h]
    \centering
    \includegraphics[width=.89\linewidth]{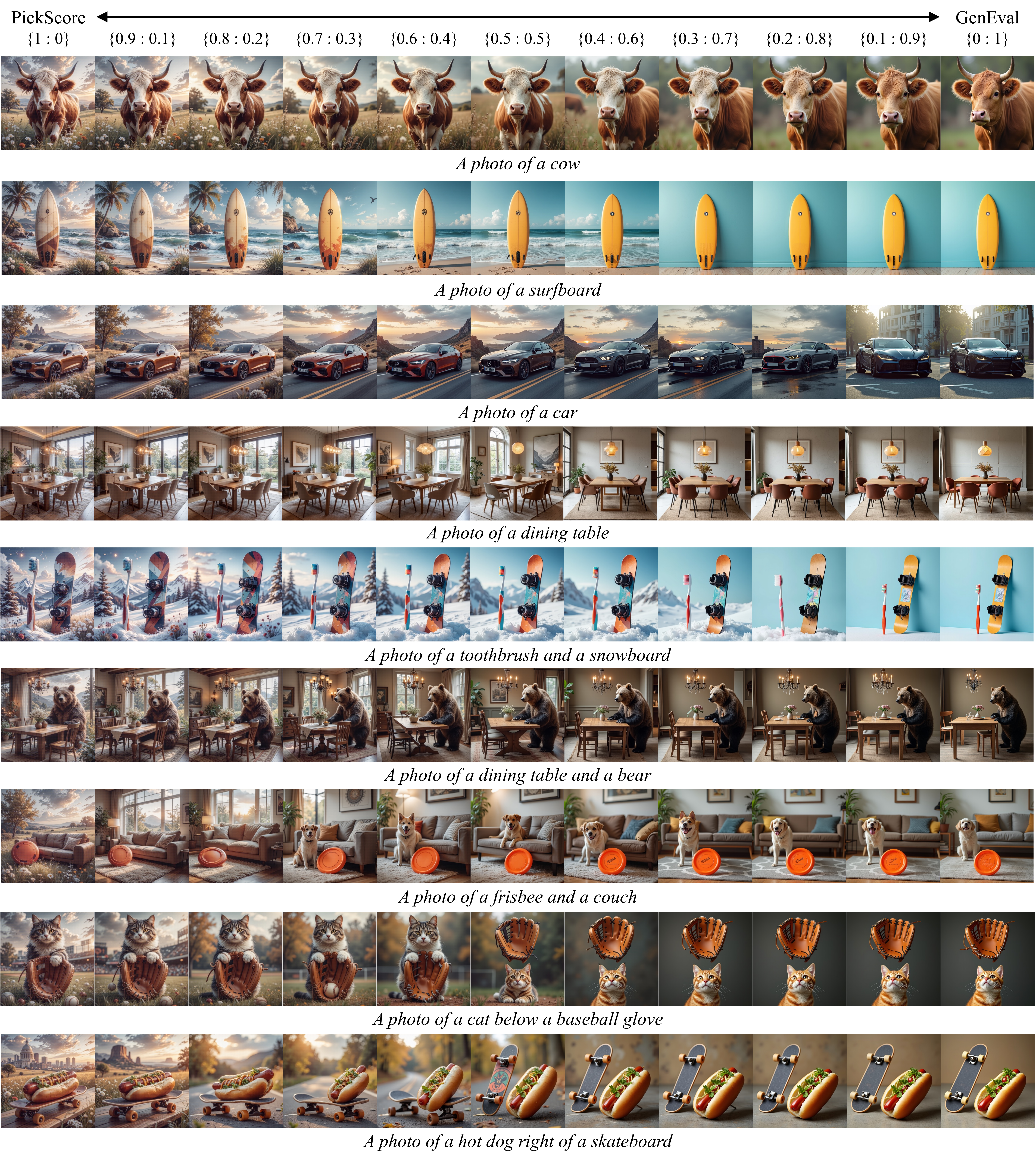}
    \vspace{-.3cm}
     \caption{Visual comparison across different merging ratios between PickScore and GenEval.}  
     \label{fig:flux-m3-gp-merging}
\end{figure*}

\begin{figure*}[h]
    \centering
    \includegraphics[width=1.0\linewidth]{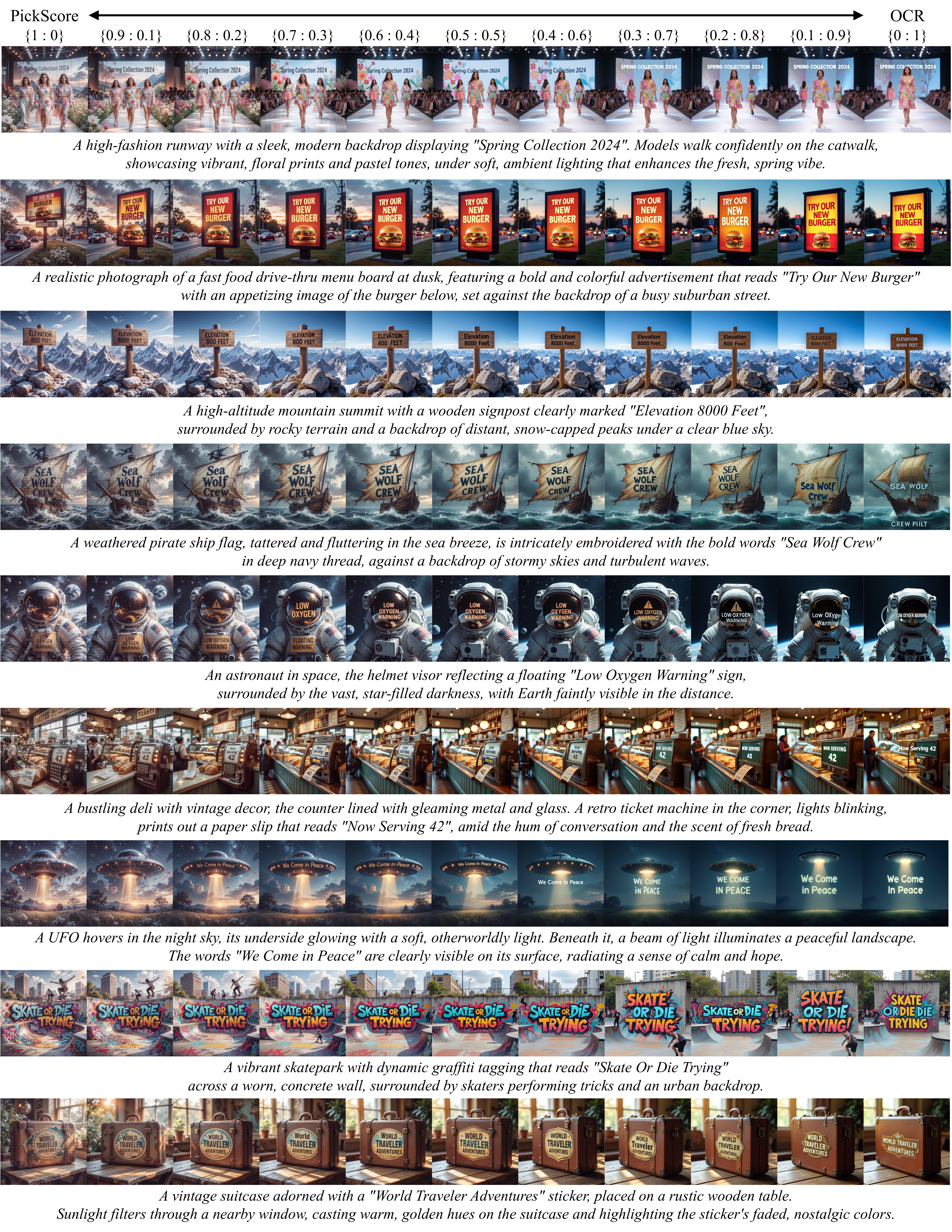}
    \vspace{-.7cm}
     \caption{Visual comparison across different merging ratios between PickScore and OCR.}  
     \label{fig:flux-m3-po-merging}
     \vspace{-.3cm}
\end{figure*}

\begin{figure*}[h]
    \centering
    \includegraphics[width=1.0\linewidth]{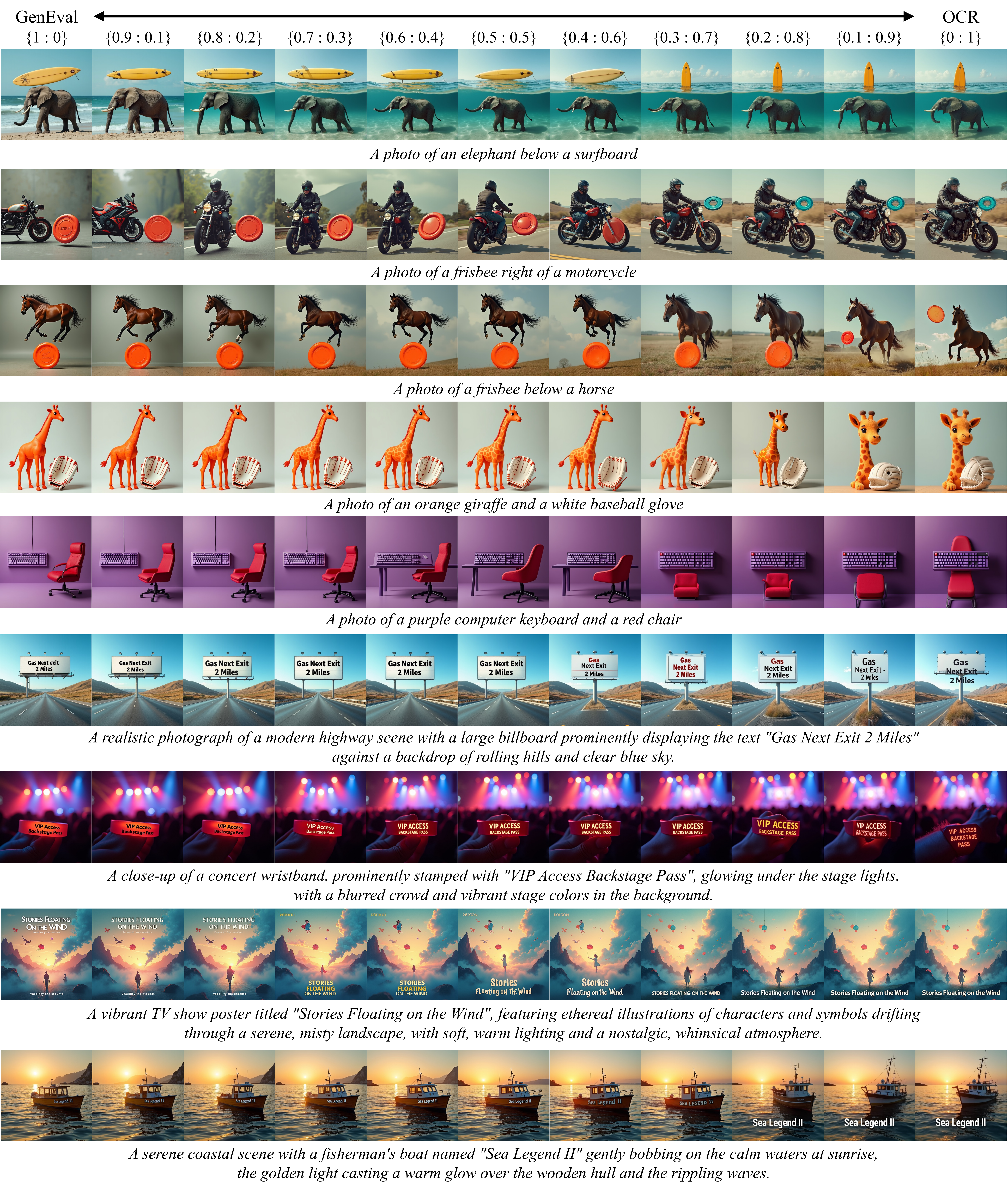}
    \vspace{-.7cm}
     \caption{Visual comparison across different merging ratios between GenEval and OCR.}  
     \label{fig:flux-m3-go-merging}
     \vspace{-.3cm}
\end{figure*}

\begin{figure*}[h]
    \centering
    \includegraphics[width=.9\linewidth]{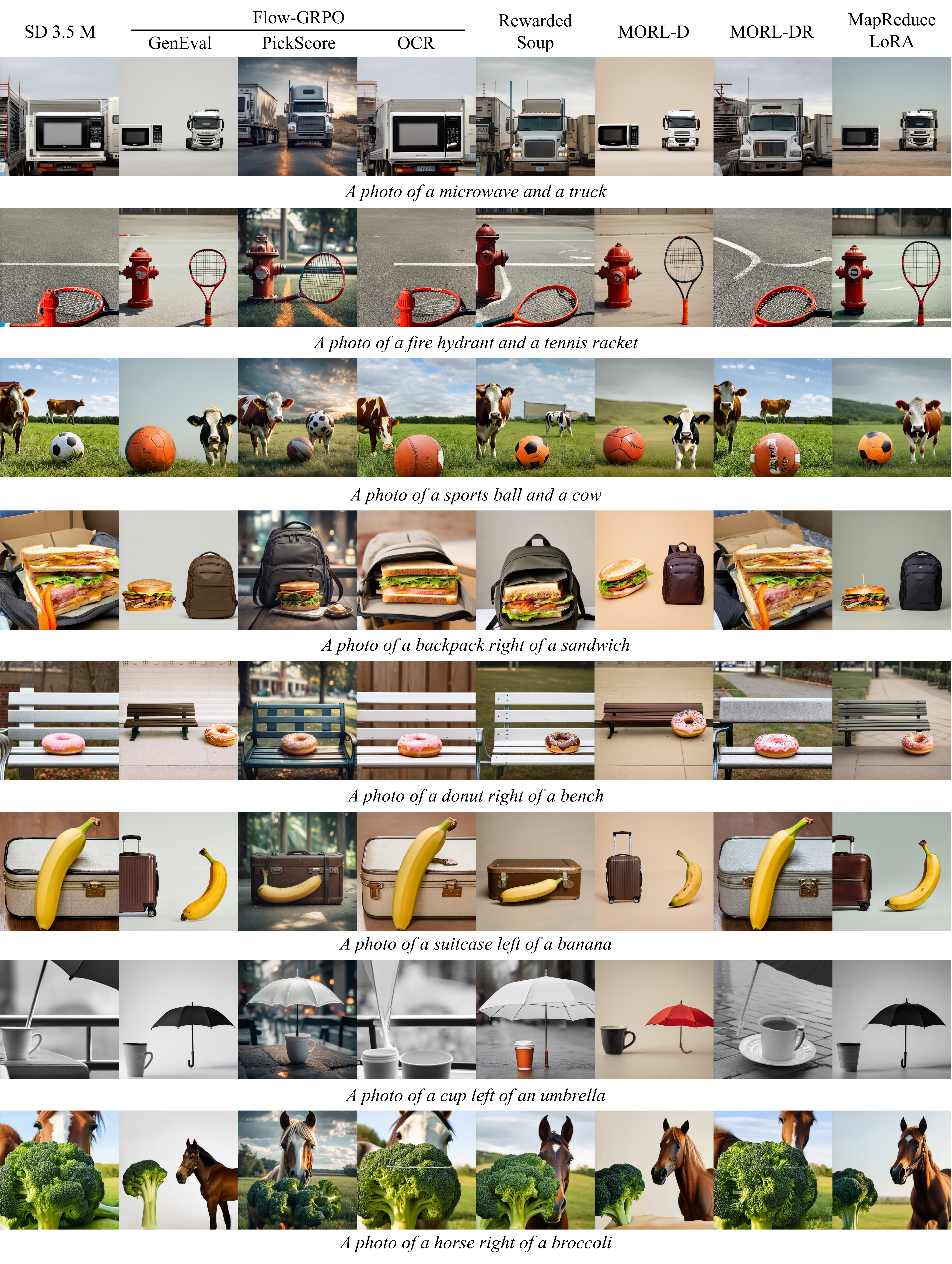}
     \caption{Visual comparison across all methods.}  
     \label{fig:sd35m-comp-all-geneval}
\end{figure*}

\begin{figure*}[h]
    \centering
    \includegraphics[width=.9\linewidth]{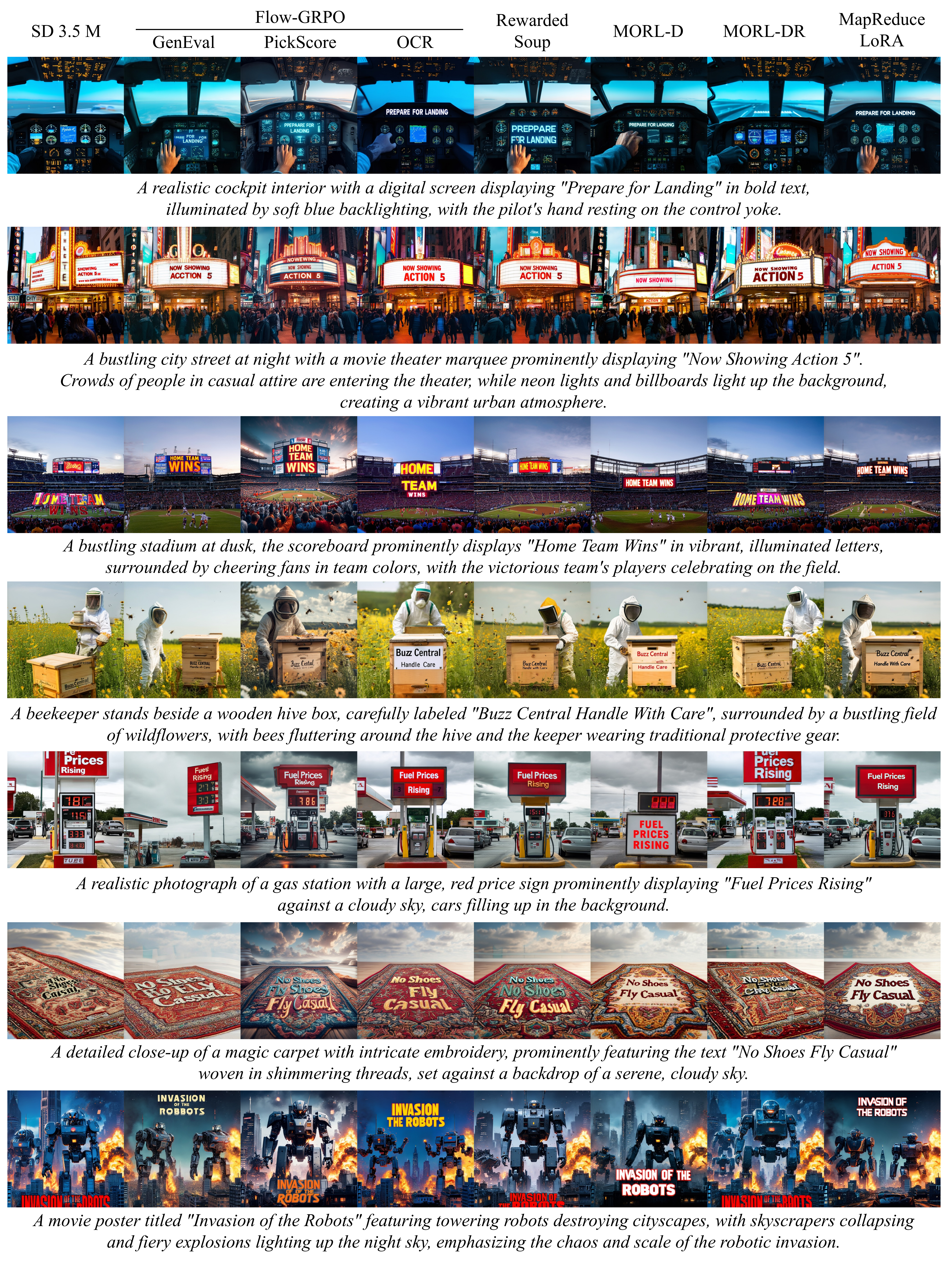}
     \caption{Visual comparison across all methods on text rendering quality.}  
     \label{fig:sd35m-comp-all-ocr}
\end{figure*}

Figs~\ref{fig:flux-m3-gp-merging}, \ref{fig:flux-m3-po-merging} and \ref{fig:flux-m3-go-merging} demonstrate the visual comparison of two rewards with different merging ratios. Figs~\ref{fig:sd35m-comp-all-geneval} and \ref{fig:sd35m-comp-all-ocr} demonstrate the visual comparison of all methods, including Flow-GRPO~\cite{flow_grpo_arxiv'25} tuned on GenEval~\cite{GenEval_NeurIPS'23}, PickScore~\cite{PickScore_NeurIPS'23}, OCR~\cite{paddleocr_arxiv'25}, Rewarded Soup~\cite{reward_soup_nips'23}, MORL-D, MORL-DR and MapReduce LoRA. PickScore and OCR guidance cause strong reward overfitting: the former drives the model toward a narrow, high-scoring aesthetic style, while the latter encourages increasingly large and prominent text regardless of overall visual harmony or contextual appropriateness. Our proposed MapReduce LoRA mitigates these issues by iteratively merging per-reward experts, enabling the model to discover a more balanced (and often near-optimal) preference that harmonizes aesthetic fidelity with text readability.

\clearpage
\subsection{Text-to-Video qualitative results}
\label{supp:t2v_more_qual}
\begin{figure*}[h]
    \centering
    \includegraphics[width=1.0\linewidth]{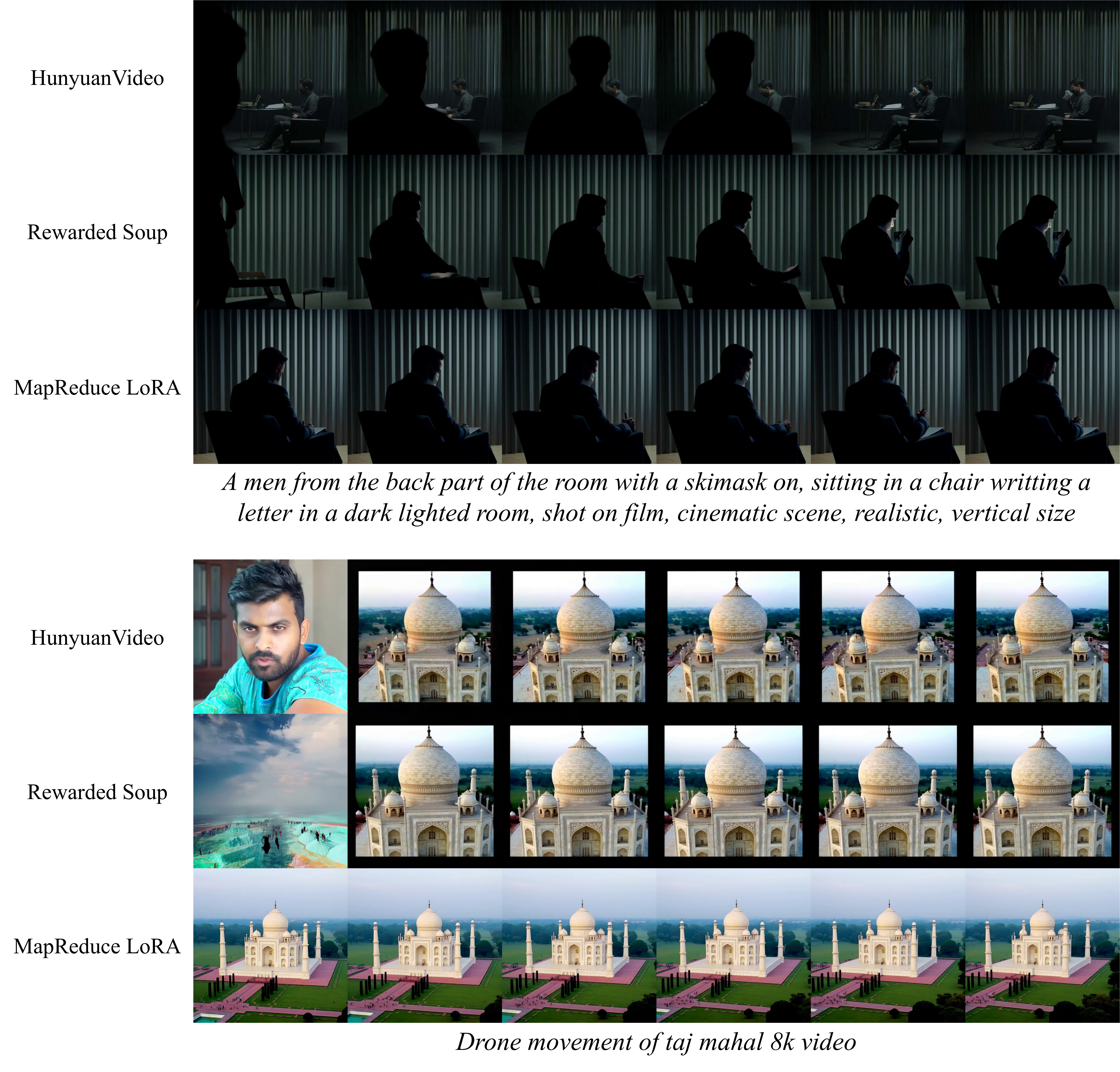}
    \vspace{-.7cm}
     \caption{\textbf{Visual comparison of HunyuanVideo~\cite{hunyuanvideo_arxiv'25}, Rewarded Soup~\cite{reward_soup_nips'23}, and MapReduce LoRA (Ours).} In the upper case, MapReduce LoRA more faithfully renders the intended motion and writing action described in the prompt. In the lower case, MapReduce LoRA better adheres to the specified drone-movement control, producing a trajectory that aligns with the prompt.}  
     \label{fig:t2v-rs-comp}
     \vspace{-.3cm}
\end{figure*}

\begin{figure*}[h]
    \centering
    \includegraphics[width=1.0\linewidth]{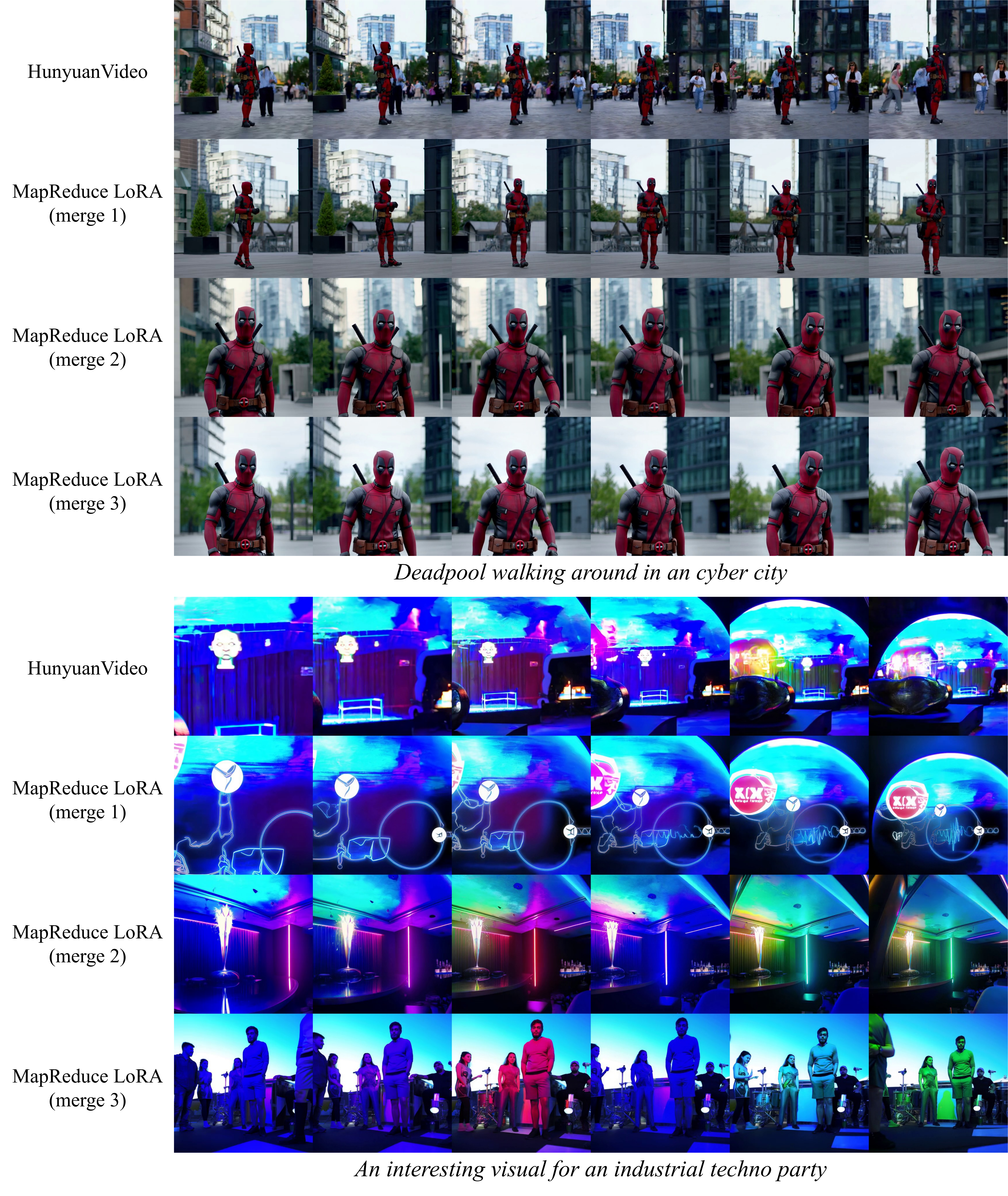}
    \vspace{-.7cm}
     \caption{\textbf{Visual performance across two cases under different MapReduce LoRA merging iterations.} In the upper case, the HunyuanVideo result fails to depict the walking motion, while increasing the merging iterations progressively restores natural walking dynamics and improves the background building quality. In the lower case, the initial generation shows limited visual fidelity, whereas successive merging iterations consistently enhance the overall clarity and scene quality.}  
     \label{fig:t2v-merging-comp-1}
     \vspace{-.3cm}
\end{figure*}

\begin{figure*}[h]
    \centering
    \includegraphics[width=1.0\linewidth]{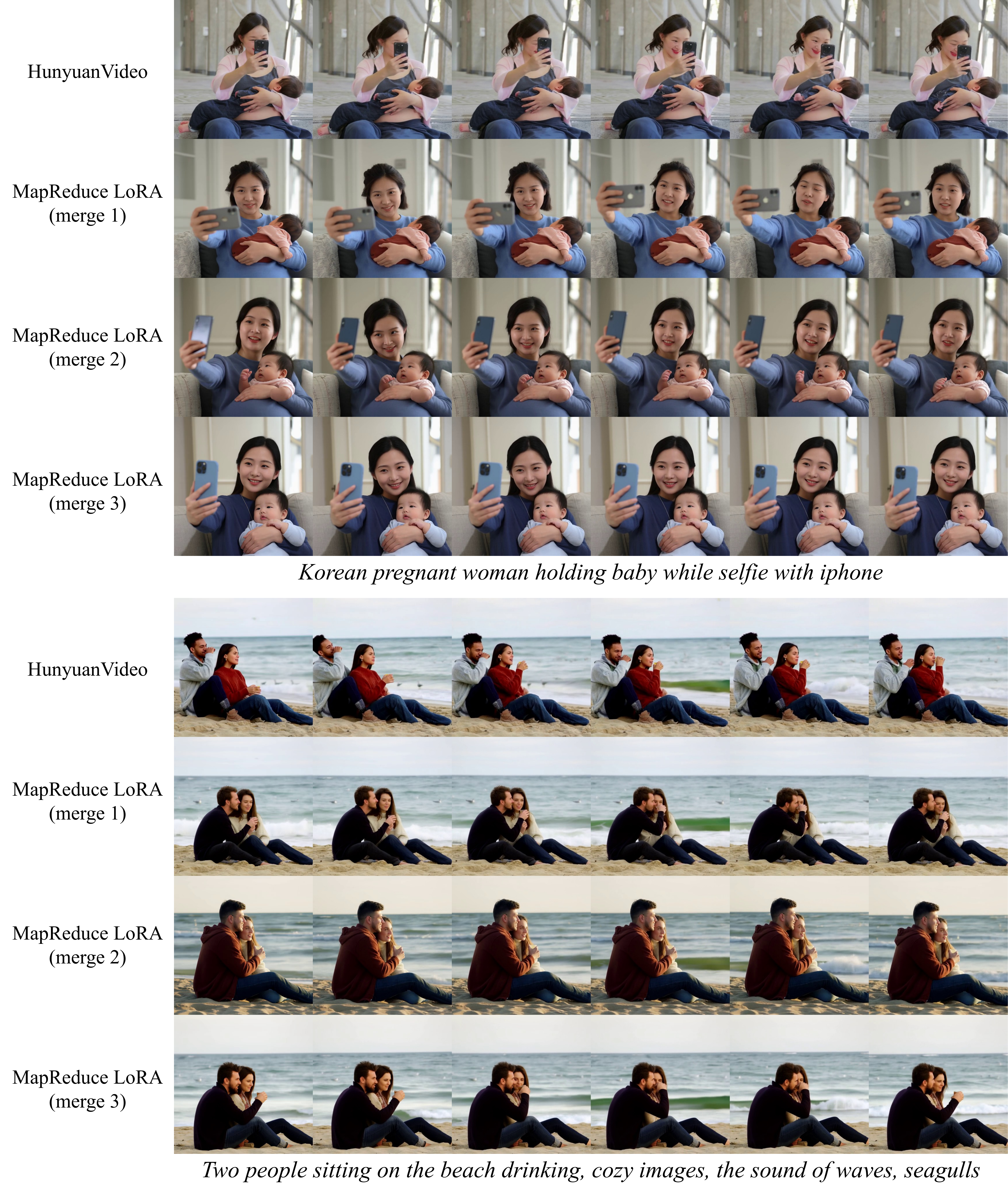}
    \vspace{-.7cm}
     \caption{\textbf{Visual performance across two cases under different MapReduce LoRA merging iterations.} In the upper case, the HunyuanVideo result shows limited visual clarity—partly due to suboptimal motion quality—while increasing merging iterations progressively refine facial and background details, improve pose naturalness, and enhance overall realism. In the lower case, the initial generation offers limited scene clarity—the drinking containers are barely visible—whereas additional merging iterations clearly render these objects and deliver sharper, more coherent visuals.}  
     \label{fig:t2v-merging-comp-2}
     \vspace{-.3cm}
\end{figure*}

Fig.~\ref{fig:t2v-rs-comp} demonstrates the visual comparison of HunyuanVideo~\cite{hunyuanvideo_arxiv'25}, Rewarded Soup~\cite{reward_soup_nips'23} and our proposed MapReduce LoRA. Also, Figs.~\ref{fig:t2v-merging-comp-1} and \ref{fig:t2v-merging-comp-2} demonstrate the visual performance of merging progress that advances the visual quality, motion quality, and prompt alignment.

%% file: sec/supp/6_limitations.tex
\clearpage
\section{Limitations and Future Works}
\label{supp:sec:limitations}

This paper presents MapReduce LoRA, a simple, scalable recipe for systematically pushing the multi-preference Pareto front and enabling practical post-training customization. We note several scope considerations and opportunities for further study:

\begin{itemize}
    \item \textbf{Scaling to more preferences:} In Text-to-Image, we validate on five targeted preferences; extending to larger numbers of preferences is a promising scaling direction.

    \item \textbf{Merging policies and schedules:} We default to uniform averaging and compare a few merge frequencies under fixed training steps; exploring adaptive/learned policies and schedules may yield further gains.

    \item \textbf{\textit{Architecture-agnostic} Reward-aware Token Embedding (RaTE):} RaTE is lightweight and effective on Stable Diffusion series models, which contain explicit cross-attention between text and image information, but is less reliable for joint sequence models, \ie, FLUX. Exploring model-agnostic designs is a practical direction.

\end{itemize}
We leave these promising directions for future work.

%% file: main.bib
@String(CVPR= {Proceedings of the IEEE/CVF Conference on Computer Vision and Pattern Recognition (CVPR)})

@String(ICCV= {Proceedings of the International Conference on Computer Vision (ICCV)})

@String(ECCV= {Proceedings of the European Conference on Computer Vision (ECCV)})

@String(NeurIPS= {Advances in Neural Information Processing Systems (NeurIPS)})

@String(ICML = {Proceedings of the International Conference on Machine Learning (ICML)})

@String(ICLR = {Proceedings of the International Conference on Learning Representations (ICLR)})

@String(ACL= {Proceedings of the Association for Computational Linguistics (ACL)})

@String(TMLR = {Transactions on Machine Learning Research (TMLR)})

@String(arXiv = {arXiv preprint})

@inproceedings{CAT_NeurIPS24,
  author       = {Chieh{-}Yun Chen and
                  Chiang Tseng and
                  Li{-}Wu Tsao and
                  Hong{-}Han Shuai},
  editor       = {Amir Globersons and
                  Lester Mackey and
                  Danielle Belgrave and
                  Angela Fan and
                  Ulrich Paquet and
                  Jakub M. Tomczak and
                  Cheng Zhang},
  title        = {A Cat Is {A} Cat (Not {A} Dog!): Unraveling Information Mix-ups in
                  Text-to-Image Encoders through Causal Analysis and Embedding Optimization},
  booktitle    = NeurIPS,
  year         = {2024},
  @url          = {http://papers.nips.cc/paper\_files/paper/2024/hash/6a69d44b3386e50c06f7107ef4f29302-Abstract-Conference.html},
  @timestamp    = {Thu, 13 Feb 2025 16:56:43 +0100},
  @biburl       = {https://dblp.org/rec/conf/nips/ChenTTS24.bib},
  @bibsource    = {dblp computer science bibliography, https://dblp.org}
}

@article{fedavg_pami23,
  author       = {Tao Sun and
                  Dongsheng Li and
                  Bao Wang},
  title        = {Decentralized Federated Averaging},
  journal      = {{IEEE} Trans. Pattern Anal. Mach. Intell.},
  year         = {2023},
  url          = {https://doi.org/10.1109/TPAMI.2022.3196503},
  doi          = {10.1109/TPAMI.2022.3196503},
  bibsource    = {dblp computer science bibliography, https://dblp.org}
}

@article{diloco_arxiv23,
  author       = {Arthur Douillard and
                  Qixuang Feng and
                  Andrei A. Rusu and
                  Rachita Chhaparia and
                  Yani Donchev and
                  Adhiguna Kuncoro and
                  Marc'Aurelio Ranzato and
                  Arthur Szlam and
                  Jiajun Shen},
  title        = {DiLoCo: Distributed Low-Communication Training of Language Models},
  journal      = {arXiv preprint arXiv:2311.08105},
  year         = {2023},
}

@article{wan_arxiv25,
  author       = {Ang Wang and others},
  title        = {Wan: Open and Advanced Large-Scale Video Generative Models},
  journal      = {arXiv preprint arXiv:2503.20314},
  year         = {2025},
}

@article{bagel_arxiv25,
  author       = {Chaorui Deng and
                  Deyao Zhu and
                  Kunchang Li and
                  Chenhui Gou and
                  Feng Li and
                  Zeyu Wang and
                  Shu Zhong and
                  Weihao Yu and
                  Xiaonan Nie and
                  Ziang Song and
                  Shi Guang and
                  Haoqi Fan},
  title        = {Emerging Properties in Unified Multimodal Pretraining},
  journal      = {arXiv preprint arXiv:2505.14683},
  year         = {2025},
}

@article{moviegen_arxiv24,
  author       = {Adam Polyak and others},
  title        = {Movie {G}en: {A} Cast of Media Foundation Models},
  journal      = {arXiv preprint arXiv:2410.13720},
  year         = {2024},
}

@inproceedings{dpo_neurips23,
  author       = {Rafael Rafailov and
                  Archit Sharma and
                  Eric Mitchell and
                  Christopher D. Manning and
                  Stefano Ermon and
                  Chelsea Finn},
  editor       = {Alice Oh and
                  Tristan Naumann and
                  Amir Globerson and
                  Kate Saenko and
                  Moritz Hardt and
                  Sergey Levine},
  title        = {Direct Preference Optimization: Your Language Model is Secretly a
                  Reward Model},
  booktitle    = NeurIPS,
  year         = {2023},
  url          = {http://papers.nips.cc/paper\_files/paper/2023/hash/a85b405ed65c6477a4fe8302b5e06ce7-Abstract-Conference.html},
  bibsource    = {dblp computer science bibliography, https://dblp.org}
}

@article{ppo_arxiv17,
  author       = {John Schulman and
                  Filip Wolski and
                  Prafulla Dhariwal and
                  Alec Radford and
                  Oleg Klimov},
  title        = {Proximal Policy Optimization Algorithms},
  journal      = {arXiv preprint arXiv:1707.06347},
  year         = {2017},
}

@inproceedings{instructgpt_neurips22,
  author       = {Long Ouyang and others},
  editor       = {Sanmi Koyejo and
                  S. Mohamed and
                  A. Agarwal and
                  Danielle Belgrave and
                  K. Cho and
                  A. Oh},
  title        = {Training language models to follow instructions with human feedback},
  booktitle    = NeurIPS,
  year         = {2022},
  url          = {http://papers.nips.cc/paper\_files/paper/2022/hash/b1efde53be364a73914f58805a001731-Abstract-Conference.html},
  bibsource    = {dblp computer science bibliography, https://dblp.org}
}

@inproceedings{ddpo_iclr24,
  author       = {Kevin Black and
                  Michael Janner and
                  Yilun Du and
                  Ilya Kostrikov and
                  Sergey Levine},
  title        = {Training Diffusion Models with Reinforcement Learning},
  booktitle    = ICLR,
  year         = {2024},
  url          = {https://openreview.net/forum?id=YCWjhGrJFD},
  bibsource    = {dblp computer science bibliography, https://dblp.org}
}

@article{shf_arxiv20,
  author       = {Nisan Stiennon and
                  Long Ouyang and
                  Jeff Wu and
                  Daniel M. Ziegler and
                  Ryan Lowe and
                  Chelsea Voss and
                  Alec Radford and
                  Dario Amodei and
                  Paul F. Christiano},
  title        = {Learning to summarize from human feedback},
  journal      = {arXiv preprint arXiv:2009.01325},
  year         = {2020},
}

@inproceedings{atari_rlhf_neurIPS18,
  author       = {Borja Ibarz and
                  Jan Leike and
                  Tobias Pohlen and
                  Geoffrey Irving and
                  Shane Legg and
                  Dario Amodei},
  editor       = {Samy Bengio and
                  Hanna M. Wallach and
                  Hugo Larochelle and
                  Kristen Grauman and
                  Nicol{\`{o}} Cesa{-}Bianchi and
                  Roman Garnett},
  title        = {Reward learning from human preferences and demonstrations in Atari},
  booktitle    = NeurIPS,
  year         = {2018},
  url          = {https://proceedings.neurips.cc/paper/2018/hash/8cbe9ce23f42628c98f80fa0fac8b19a-Abstract.html},
  bibsource    = {dblp computer science bibliography, https://dblp.org}
}

@inproceedings{rlhf_neurips17,
  author       = {Paul F. Christiano and
                  Jan Leike and
                  Tom B. Brown and
                  Miljan Martic and
                  Shane Legg and
                  Dario Amodei},
  editor       = {Isabelle Guyon and
                  Ulrike von Luxburg and
                  Samy Bengio and
                  Hanna M. Wallach and
                  Rob Fergus and
                  S. V. N. Vishwanathan and
                  Roman Garnett},
  title        = {Deep Reinforcement Learning from Human Preferences},
  booktitle    = NeurIPS,
  year         = {2017},
  url          = {https://proceedings.neurips.cc/paper/2017/hash/d5e2c0adad503c91f91df240d0cd4e49-Abstract.html},
  bibsource    = {dblp computer science bibliography, https://dblp.org}
}

@article{MOPO_arxiv25,
  author       = {Akhil Agnihotri and
                  Rahul Jain and
                  Deepak Ramachandran and
                  Zheng Wen},
  title        = {Multi-Objective Preference Optimization: Improving Human Alignment
                  of Generative Models},
  journal      = {arXiv preprint arXiv:2505.10892},
  year         = {2025},
}

@inproceedings{selma_neurips24,
  author       = {Jialu Li and
                  Jaemin Cho and
                  Yi{-}Lin Sung and
                  Jaehong Yoon and
                  Mohit Bansal},
  editor       = {Amir Globersons and
                  Lester Mackey and
                  Danielle Belgrave and
                  Angela Fan and
                  Ulrich Paquet and
                  Jakub M. Tomczak and
                  Cheng Zhang},
  title        = {{SELMA:} Learning and Merging Skill-Specific Text-to-Image Experts
                  with Auto-Generated Data},
  booktitle    = NeurIPS,
  year         = {2024},
  url          = {http://papers.nips.cc/paper\_files/paper/2024/hash/43d8e5fc816c692f342493331d5e98fc-Abstract-Conference.html},
  bibsource    = {dblp computer science bibliography, https://dblp.org}
}

@inproceedings{bonesoup_acl25,
  author       = {Guofu Xie and
                  Xiao Zhang and
                  Ting Yao and
                  Yunsheng Shi},
  editor       = {Wanxiang Che and
                  Joyce Nabende and
                  Ekaterina Shutova and
                  Mohammad Taher Pilehvar},
  title        = {Bone Soups: {A} Seek-and-Soup Model Merging Approach for Controllable
                  Multi-Objective Generation},
  booktitle    = ACL,
  @pages        = {27237--27263},
  @publisher    = {Association for Computational Linguistics},
  year         = {2025},
  url          = {https://aclanthology.org/2025.acl-long.1322/},
  timestamp    = {Sun, 02 Nov 2025 21:27:24 +0100},
  biburl       = {https://dblp.org/rec/conf/acl/XieZYS25.bib},
  bibsource    = {dblp computer science bibliography, https://dblp.org}
}

@article{multilora_tmlr24,
  author       = {Ming Zhong and
                  Yelong Shen and
                  Shuohang Wang and
                  Yadong Lu and
                  Yizhu Jiao and
                  Siru Ouyang and
                  Donghan Yu and
                  Jiawei Han and
                  Weizhu Chen},
  title        = {Multi-LoRA Composition for Image Generation},
  journal      = TMLR,
  year         = {2024},
  url          = {https://openreview.net/forum?id=25FT0DqhVZ},
  bibsource    = {dblp computer science bibliography, https://dblp.org}
}

@inproceedings{rf_iclr23,
  author       = {Xingchao Liu and
                  Chengyue Gong and
                  Qiang Liu},
  title        = {Flow Straight and Fast: Learning to Generate and Transfer Data with
                  Rectified Flow},
  booktitle    = ICLR,
  year         = {2023},
  url          = {https://openreview.net/forum?id=XVjTT1nw5z},
  bibsource    = {dblp computer science bibliography, https://dblp.org}
}

@book{paretofront_1896,
  author    = {Vilfredo Pareto},
  title     = {Cours d'{\'e}conomie politique},
  year      = {1896},
  publisher = {F. Rouge},
}

@article{schulman2025lora,
  author = {John Schulman and Thinking Machines Lab},
  title = {LoRA Without Regret},
  journal = {Thinking Machines Lab: Connectionism},
  year = {2025},
  note = {\url{https://thinkingmachines.ai/blog/lora/}},
  doi = {10.64434/tml.20250929}
}

@inproceedings{sd3_icml24,
  author       = {Patrick Esser and
                  others},
  title        = {Scaling Rectified Flow Transformers for High-Resolution Image Synthesis},
  booktitle    = ICML,
  year         = {2024},
  url          = {https://openreview.net/forum?id=FPnUhsQJ5B},
  bibsource    = {dblp computer science bibliography, https://dblp.org}
}

@article{dancegrpo_arxiv25,
  author       = {Zeyue Xue and
                  Jie Wu and
                  Yu Gao and
                  Fangyuan Kong and
                  Lingting Zhu and
                  Mengzhao Chen and
                  Zhiheng Liu and
                  Wei Liu and
                  Qiushan Guo and
                  Weilin Huang and
                  Ping Luo},
  title        = {Dance{GRPO}: Unleashing {GRPO} on Visual Generation},
  journal      = {arXiv preprint arXiv:2505.07818},
  year         = {2025},
}

@article{videoalign_arxiv25,
  author       = {Jie Liu and
                  Gongye Liu and
                  Jiajun Liang and
                  Ziyang Yuan and
                  Xiaokun Liu and
                  Mingwu Zheng and
                  Xiele Wu and
                  Qiulin Wang and
                  Wenyu Qin and
                  Menghan Xia and
                  Xintao Wang and
                  Xiaohong Liu and
                  Fei Yang and
                  Pengfei Wan and
                  Di Zhang and
                  Kun Gai and
                  Yujiu Yang and
                  Wanli Ouyang},
  title        = {Improving Video Generation with Human Feedback},
  journal      = {arXiv preprint arXiv:2501.13918},
  year         = {2025},
}

@misc{sd3m_hf,
  title={Stable Diffusion 3 Medium},
  author={Stability AI},
  howpublished={\url{https://huggingface.co/stabilityai/stable-diffusion-3-medium}},
  year={2024}
}

@misc{sd35m_hf,
  title={Stable Diffusion 3.5 Medium},
  author={Stability AI},
  howpublished={\url{https://huggingface.co/stabilityai/stable-diffusion-3.5-medium}},
  year={2024}
}

@misc{flux24,
    author={Black Forest Labs},
    title={{FLUX}},
    year={2024},
    howpublished={\url{https://github.com/black-forest-labs/flux}},
}

@inproceedings{
GenAIBench_CVPRW24,
title={Gen{AI}-Bench: A Holistic Benchmark for Compositional Text-to-Visual Generation},
author={Baiqi Li and Zhiqiu Lin and Deepak Pathak and Jiayao Emily Li and Xide Xia and Graham Neubig and Pengchuan Zhang and Deva Ramanan},
booktitle={Synthetic Data for Computer Vision Workshop @ CVPR},
year={2024},
url={https://openreview.net/forum?id=hJm7qnW3ym}
}

@article{partiprompt_TMLR22,
  author       = {Jiahui Yu and
                  Yuanzhong Xu and
                  Jing Yu Koh and
                  Thang Luong and
                  Gunjan Baid and
                  Zirui Wang and
                  Vijay Vasudevan and
                  Alexander Ku and
                  others},
  title        = {Scaling Autoregressive Models for Content-Rich Text-to-Image Generation},
  journal      = TMLR,
  year         = {2022},
  url          = {https://openreview.net/forum?id=AFDcYJKhND},
  bibsource    = {dblp computer science bibliography, https://dblp.org}
}

@inproceedings{VQAScore_ECCV24,
  author       = {Zhiqiu Lin and
                  Deepak Pathak and
                  Baiqi Li and
                  Jiayao Li and
                  Xide Xia and
                  Graham Neubig and
                  Pengchuan Zhang and
                  Deva Ramanan},
  @editor       = {Ales Leonardis and
                  Elisa Ricci and
                  Stefan Rxoth and
                  Olga Russakovsky and
                  Torsten Sattler and
                  G{\"{u}}l Varol},
  title        = {Evaluating Text-to-Visual Generation with Image-to-Text Generation},
  booktitle    = ECCV,
  @series       = {Lecture Notes in Computer Science},
  @volume       = {15067},
  @publisher    = {Springer},
  year         = {2024},
  @url          = {https://doi.org/10.1007/978-3-031-72673-6\_20},
  @doi          = {10.1007/978-3-031-72673-6\_20},
  @timestamp    = {Wed, 06 Nov 2024 22:17:13 +0100},
  @biburl       = {https://dblp.org/rec/conf/eccv/LinPLLXNZR24.bib},
  @bibsource    = {dblp computer science bibliography, https://dblp.org}
}

@InProceedings{Chen_2025_ICCV,
    author    = {Chen, Chieh-Yun and Shi, Min and Zhang, Gong and Shi, Humphrey},
    title     = {T2I-Copilot: A Training-Free Multi-Agent Text-to-Image System for Enhanced Prompt Interpretation and Interactive Generation},
    booktitle = {Proceedings of the IEEE/CVF International Conference on Computer Vision (ICCV)},
    month     = {October},
    year      = {2025},
}

@article{llama2_arxiv23,
  author       = {Hugo Touvron and others},
  title        = {Llama 2: Open Foundation and Fine-Tuned Chat Models},
  journal      = {arXiv preprint arXiv:2307.09288},
  year         = {2023},
}
